\documentclass[10pt,journal,compsoc]{IEEEtran}
\usepackage[pdftex]{graphicx}
\usepackage{amsmath}
\usepackage{amssymb}
\usepackage{bm}
\usepackage{color}
\usepackage{colortbl}
\usepackage{xcolor}
\usepackage{array}

\usepackage{algorithm,algpseudocode}

\usepackage{multirow}
\usepackage{subfigure}
\usepackage{makecell}
\usepackage{booktabs}
\usepackage{xspace}
\usepackage{gensymb}
\usepackage[colorlinks,linkcolor=black]{hyperref}

\ifCLASSOPTIONcompsoc
  \usepackage[nocompress]{cite}
\else
  \usepackage{cite}
\fi

\begin{document}
\title{3D-Aware Adversarial Makeup Generation \\for Facial Privacy Protection}
%
%
%
%

\author{Yueming Lyu,
        Yue Jiang, 
        Ziwen He,
        Bo Peng,~\IEEEmembership{Member,~IEEE},\\
		Yunfan Liu,
        and Jing Dong,~\IEEEmembership{Senior Member,~IEEE}
\IEEEcompsocitemizethanks{\IEEEcompsocthanksitem Yueming Lyu, Yue Jiang, Ziwen He, Bo Peng, Yunfan Liu, Jing Dong~(corresponding author) are with the Center for Research on Intelligent Perception and Computing~(CRIPAC), State Key Laboratory of Multimodal Artificial Intelligence Systems, Institute of Automation, Chinese Academy of Sciences~(CASIA), Beijing 100190, China, and also with the School of Artificial Intelligence, University of Chinese Academy of Sciences, Beijing 100049, China (E-mail: yueming.lv@cripac.ia.ac.cn; jiangyue2021@ia.ac.cn; ziwen.he@cripac.ia.ac.cn; bo.peng@nlpr.ia.ac.cn; yunfan.liu@cripac.ia.ac.cn; jdong@nlpr.ia.ac.cn).\protect}}


%
%

\markboth{Journal of \LaTeX\ Class Files,~Vol.~14, No.~8, August~2015}%
{Shell \MakeLowercase{\textit{et al.}}: Bare Advanced Demo of IEEEtran.cls for IEEE Computer Society Journals}
%



\IEEEtitleabstractindextext{%
\begin{abstract}
The privacy and security of face data on social media are facing unprecedented challenges as it is vulnerable to unauthorized access and identification. A common practice for solving this problem is to modify the original data so that it could be protected from being recognized by malicious face recognition (FR) systems. However, such ``adversarial examples'' obtained by existing methods usually suffer from low transferability and poor image quality, which severely limits the application of these methods in real-world scenarios. In this paper, we propose a 3D-Aware Adversarial Makeup Generation GAN (3DAM-GAN). which aims to improve the quality and transferability of synthetic makeup for identity information concealing. Specifically, a UV-based generator consisting of a novel Makeup Adjustment Module (MAM) and Makeup Transfer Module (MTM) is designed to render realistic and robust makeup with the aid of symmetric characteristics of human faces. Moreover, a makeup attack mechanism with an ensemble training strategy is proposed to boost the transferability of black-box models. Extensive experiment results on several benchmark datasets demonstrate that 3DAM-GAN could effectively protect faces against various FR models, including both publicly available state-of-the-art models and commercial face verification APIs, such as Face++, Baidu and Aliyun.
\end{abstract}

}
\maketitle

\IEEEdisplaynontitleabstractindextext

%
\IEEEpeerreviewmaketitle

\ifCLASSOPTIONcompsoc
\IEEEraisesectionheading{\section{Introduction}\label{sec:introduction}}
\else
\section{Introduction}
\label{sec:introduction}
\fi

\IEEEPARstart{T}{he} booming development of mobile devices and the Internet has made it easier than ever to capture and share portrait images of individuals online. However, such massive face data being uploaded to various online social media platforms, such as Facebook, Twitter and TikTok, could be easily identified, processed and utilized for malicious purposes. For example, users could be recognized by unauthorized face recognition (FR) systems and their profiles could be analyzed without consent, which imposes a considerable threat to the privacy and security of personal information. Therefore, protecting the privacy of individuals from being unconsciously identified and exposed to unauthorized FR systems is an urgent problem to be solved.

\begin{figure}[tp]
	\centering
		\includegraphics[width=\columnwidth]{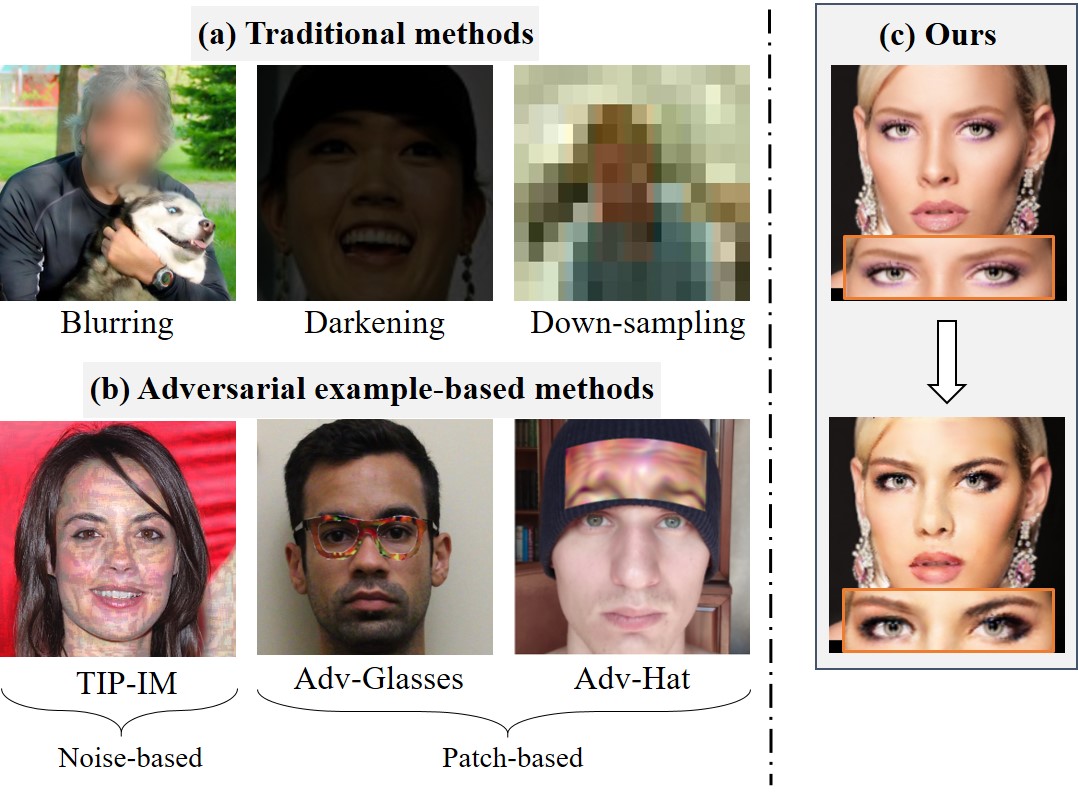}
		\caption{Comparison of facial privacy protection results generated by (a) traditional methods~\cite{yang2022study,wilber2016can,ryoo2017privacy}, (b) adversarial example-based methods~\cite{yang2021towards,sharif2016accessorize,komkov2021advhat}, and (c) our method. Results in (a) and (b)  (except for TIP-IM) are directly obtained from the referenced paper.}
	\label{fig:shouye}
\end{figure}

Traditional methods conceal the identity information in face images by various low-level operations, including blurring~\cite{yang2022study}, darkening~\cite{wilber2016can}, and down-sampling~\cite{ryoo2017privacy}. However, these methods have drawbacks, including (a) losses in image quality, (b) difficulties in recognizing other features in the image such as clothing or background, and (c) significant changes to the visual identity of the face, as shown in Fig.~\ref{fig:shouye} (a).
Furthermore, simply not uploading the image does provide true privacy protection. However, there are situations where users willingly share their photos or are required to upload them, such as on social media or for professional purposes. Another approach~\cite{zhu2021one,xu2022high,xu2022styleswap} to protecting privacy involves replacing the identity with a completely new face using methods like StyleGAN2~\cite{karras2020analyzing}. However, in certain contexts, users may prefer displaying their actual facial photos rather than completely different faces.

With the fast development of deep neural networks (DNNs), recent studies~\cite{rajabi2021practicality,cherepanova2021lowkey,yang2021towards} model the privacy protecting process as adversarial attacks to unauthorized FR models. These methods mainly apply noise-based or patch-based perturbations to the original data (often referred to as ``adversarial example''), and thus more textural details can be preserved in the resultant face images than traditional methods, as shown in Fig.~\ref{fig:shouye} (b). 

However, such adversarial example-based methods have the following problems:
\begin{itemize}
\item \textbf{Low transferability to black-box FR models.}
Some methods~\cite{goodfellow2014explaining,brown2017adversarial,madry2017towards} assume that gradients of the target FR model are available when creating adversarial examples. However, methods built on such ``white-box assumption'' are impractical in real-world scenarios as the FR systems on social media are often inaccessible to users.

\item \textbf{Degraded image quality.}
As shown in Fig.~\ref{fig:shouye} (b), although some approaches~\cite{kurakin2017adversarial,dong2018boosting} are able to produce adversarial perturbations with much less intensity than traditional methods, either by injecting noise or attaching patches, they may still fail to preserve the visual quality of original images. 
\end{itemize}

To solve these problems, we aim to protect personal privacy by creating adversarial examples which are both visually appealing and applicable in real-world scenarios. Different from existing approaches, the key idea of our method is to produce the perturbations by transferring the makeup style of reference faces to source images. In this way, the protected face images with natural makeup are more visually plausible to human perception, which will greatly reduce the influence on user experience and simultaneously conceal the identity from unauthorized FR systems. 

In this paper, we propose a novel 3D-Aware Adversarial Makeup Generation GAN (3DAM-GAN) which aims to generate natural adversarial examples by makeup transfer. To deal with the geometric inconsistency between source and reference faces, a novel UV-generator is proposed to align facial textures from the 3D perspective. Specifically, with the aid of the inherent symmetry of human faces as well as the visibility map in UV space, a Makeup Adjustment Module (MAM) is proposed to learn robust style information of makeup by eliminating the interference of occlusion and shadow in reference images. Moreover, since texture maps in the UV space are spatially aligned (\emph{i.e.}, each position has the same semantic meaning), a Makeup Transfer Module (MTM) is also introduced to improve the accuracy of makeup transfer by learning a pixel-wise attention map. In addition, a novel makeup loss is incorporated to further improve the quality of generation results, which explicitly leverages the bilateral symmetry of faces in the UV space and provides more accurate and robust makeup supervision for the makeup transfer process. Finally, we introduce a targeted makeup attack mechanism to prevent the generated images with adversarial makeup styles from being recognized by unauthorized FR systems. Based on introducing an ensemble training strategy, 3DAM-GAN is able to find a common and informative gradient signal against various FR models, and has a greater transferability to different black-box FR systems.

Overall, our proposed method addresses the limitations of existing approaches by considering both visual quality and real-world applicability. It enables individuals to protect their privacy while maintaining a natural and visually appealing appearance in scenarios where uploading photos is necessary or desired.

We summarize our contributions as follows:
\begin{itemize}
\item We propose a novel 3D-Aware Adversarial Makeup Generation GAN (3DAM-GAN) to derive adversarial makeup from arbitrary reference images. It is able to render natural adversarial examples to protect face images from unauthorized FR systems under a more practical black-box attack setting.

\item We design a UV generator consisting of a Makeup Adjustment Module (MAM) and a Makeup Transfer Module (MTM), which help improve the visual fidelity of the synthesized makeup features. In addition, a UV makeup loss is proposed by leveraging the bilateral symmetry of faces in the UV space to further provide more accurate and robust makeup supervision. 

\item We introduce a makeup attack mechanism against various FR models with an ensemble training strategy. It helps conceal the real identity in makeup generation results from unauthorized FR systems and improve the transferability to black-box setting.

\item Extensive qualitative and quantitative experiments demonstrate that 3DAM-GAN could generate realistic adversarial examples with high success rate of identity protection.
\end{itemize}

This work is an extension of our previous work, SOGAN~\cite{lyu2021sogan}, and we have made the following improvements:
(1) the original makeup transfer method is extended to the adversarial attack setting, which could serve as a powerful solution to the facial privacy protection problem.
(2) a 3D visibility map is integrated to the Makeup Adjustment Module (MAM) to introduce prior knowledge of facial geometry, which helps deal with the shadow and occlusion in reference images, and improves the visual quality of the generated makeup. 
(3) a novel UV makeup loss is involved to further supervise the makeup transfer process more effectively. 
(4) an ensemble training strategy is proposed for solving a common and informative gradient signal against various FR models and boosting the black-box transferability. 
(5) extensive experiments have been conducted on several benchmark datasets, and the results demonstrate the effectiveness of 3DAM-GAN in protecting faces against various FR models, including commercial face verification APIs of Face++\footnote[2]{\url{https://www.faceplusplus.com/}}, Baidu\footnote[3]{\url{https://ai.baidu.com/tech/face/}} and Aliyun\footnote[4]{\url{https://vision.aliyun.com/}}. 

\section{Related work}
\label{sec:relate}
\subsection{Adversarial Attacks on Face Recognition}
Existing face recognition (FR) systems are usually built on deep neural networks, which are known to be vulnerable to adversarial examples~\cite{szegedy2013intriguing,goodfellow2014explaining,dong2019efficient,xiao2021improving}. In general, the attacks against FR systems could be divided into two categories, white-box attacks~\cite{goodfellow2014explaining,hu2021advhash,madry2017towards}, where the full access to target FR models is required to generate adversarial examples, and black-box attacks~\cite{dong2018boosting,xiao2021improving,yang2021towards,zhong2020towards}, where the internal mechanism of target FR models are completely unknown to the adversaries. Particularly, in the case of protecting user-uploaded photos on social media, white-box attacks are inapplicable since the unauthorized FR systems are usually unknown.

Thus, black-box attacks are more suitable for protecting facial privacy in real-world scenarios. Most existing methods for black-box attacks~\cite{dong2018boosting,dong2019efficient} are formulated as an optimization problem. 
For example, Dong \emph{et al.}~\cite{dong2019efficient} propose an evolutionary optimization method, which requires at least 1,000 queries to the target FR systems to synthesize a realistic adversarial face. Xiao \emph{et al.}~\cite{xiao2021improving} regularize the adversarial patch by optimizing it in the latent space of a generative model.
Unfortunately, optimization-based methods are time-consuming and may suffer from low transferability due to over-fitting~\cite{xiao2021improving}. 

Recently, Adv-Makeup~\cite{yin2021adv} proposes to synthesize imperceptible eye shadows over the orbital regions on faces to achieve target attacks under a black-box setting. AMTGAN~\cite{hu2022protecting} realizes facial privacy protection via makeup transfer under a black-box setting. It first transfers the overall reference makeup onto the corresponding source face region and then introduces ensemble training with input diversity enhancement to generate adversarial examples against the target identity. However, the above approaches based on makeup attack take little consideration of the makeup quality. For example, Adv-Makeup~\cite{yin2021adv} attaches the synthesized orbital regions to the source faces directly, which yields low-quality effects and obvious intensity changes at the boundary. AMTGAN\cite{hu2022protecting} does not optimize the makeup transfer sub-network but directly uses the generator~\cite{jiang2020psgan}. We argue that the makeup transfer quality is crucial to generate natural and stable adversarial examples, since the attack information is hidden in the makeup effect. 

Therefore, different from them, we propose a 3D-aware adversarial makeup generation method to improve the makeup effect and boost the black-box attack success rate simultaneously.

\subsection{Makeup Transfer} 
Makeup transfer aims to render makeup in source faces according to the style of reference images, while keep the identity information intact. This is a challenging task as the fine-grained makeup details are required to be accurately captured and transferred. Traditional approaches utilize image pre-processing techniques, such as facial landmark detection and alignment~\cite{xu2013automatic} or reflectance manipulation~\cite{li2015simulating}, to perform makeup transfer. Recently, with the development of generative adversarial networks (GANs), significant improvement has been achieved in improving the quality of makeup transfer results. BeautyGAN \cite{li2018beautygan} introduces a dual input/output generator and a pixel-wise color histogram loss on facial components to transfer makeup details.
Based on the Glow~\cite{kingma2018glow} model, BeautyGlow \cite{chen2019beautyglow} disentangles the latent code of input images into makeup and non-makeup parts, and then re-combines them to obtain the final makeup image. 
LADN~\cite{gu2019ladn} incorporates multiple overlapping local discriminators to better transfer local details. 
PSGAN \cite{jiang2020psgan} proposes a pose and expression robust GAN model by introducing an attentive makeup morphing module.
SOGAN~\cite{lyu2021sogan} tackles the problem of undesired shadow and occlusion in the reference images. 
PSGAN++~\cite{liu2021psgan++} proposes to achieve robust and controllable makeup transfer, as well as makeup removal.

\subsection{3D Face Reconstruction}
3D face reconstruction aims to recover the 3D information of human faces from 2D projections. As a commonly used statistical model for 3D face reconstruction, 3D Morphable Models (3DMMs)~\cite{blanz1999morphable,blanz2003face,paysan20093d} learn from 3D scans of human faces and model the major variations of both the 3D shape and texture information. With the development of deep learning, CNN-based approaches have been proposed to estimate the 3D Morphable Model parameters \cite{zhu2016face,tewari2017mofa,guo2020towards}, which improve the performance of reconstruction. Using 3DMM parameters, we can obtain a series of 3D representations, such as 3D face shape, texture, and UV texture map \cite{feng2018joint,guo2020beyond}. In addition, neural renderer \cite{kim2018deep,thies2019deferred} can be used to fit 3D face models to 2D images. In this paper, we employ 3DMM to obtain the shape and texture representation and then exploit the symmetry of UV space to synthesize adversarial face images with makeup. 
\begin{figure*}
	\centering
		\includegraphics[width=0.8\textwidth]{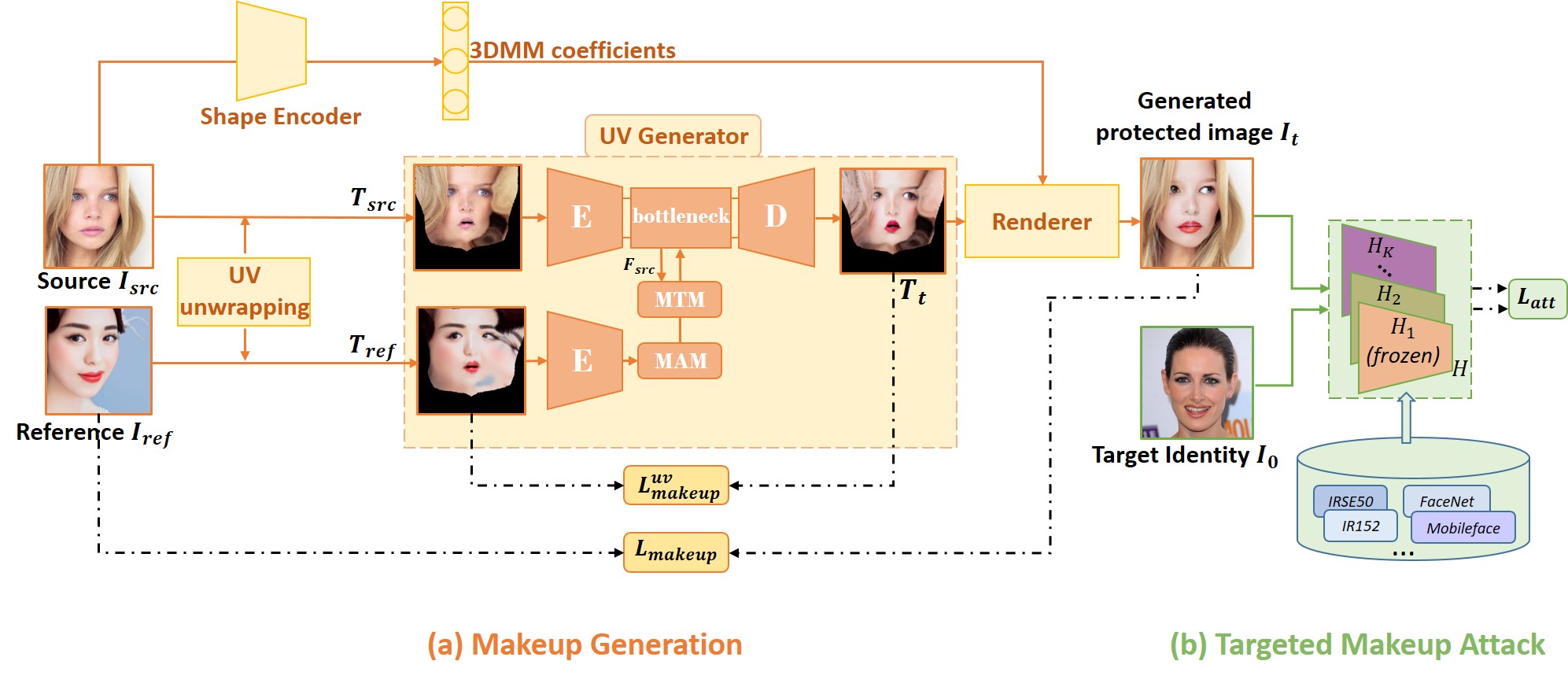}
	\caption{An overview of the framework of 3DAM-GAN. As for (a) makeup generation, we propose a UV generator to transfer the makeup in the UV space while keeping the shape unchanged. The UV generator contains two novelly designed modules, Makeup Adjustment Module (MAM) and Makeup Transfer Module (MTM), which work together to improve the makeup generation quality. Moreover, we introduce a novel UV makeup loss $\mathcal{L}_{makeup}^{uv}$ to provide more accurate makeup supervision during training. As for (b) targeted makeup attack, we consider targeted adversarial attack and perform ensemble training against pre-trained FR models $H=\{H_1,H_2,\cdots,H_K\}$ chosen from various FR systems.}
	\label{fig:pipeline}
\end{figure*}
\subsection{3D-Guided Face Synthesis}
Since 3D priors could provide semantic control on facial geometry, many methods have been proposed to integrate it into GAN-based frameworks for face image synthesis~\cite{tewari2020stylerig,deng2020disentangled,xu2020deep,hu2020face}
For example, StyleRig \cite{tewari2020stylerig} utilizes 3DMM parameters to provide rig-like controls on faces with a pre-trained StyleGAN generator with fixed weights. DiscoFaceGAN \cite{deng2020disentangled} proposes to supervise the face images generated by GANs by making them imitate the semantic of faces rendered by 3DMM with pre-defined parameters, which enables precise control of facial properties such as pose, expression, and illumination. \cite{xu2020deep} learns 3D head geometry in an unsupervised manner and achieves accurate 3D head geometry modeling and high-fidelity head pose manipulation results.
In order to improve the performance of face super-resolution, \cite{hu2020face} introduces a 3D rendering branch to obtain 3D priors of salient facial structure and identity knowledge.
In this paper, we employ 3DMM to disentangle texture and shape spaces, and apply adversarial makeup transfer in the texture space.

\section{Methodology}
\label{sec:method}

\subsection{Problem Formulation}
We propose to conceal the personal characteristics in source image $I_{src}$ by adding makeup with the style provided by reference image $I_{ref}$, so that unauthorized FR systems are not able to recover the original identity based on the generation result $I_{t} = \mathcal{F}\left(I_{src}, I_{ref}\right)$ ($\mathcal{F}$ denoted the learned mapping function). The overall pipeline of the proposed 3DAM-GAN is shown in Figure~\ref{fig:pipeline}, which mainly consists of two parts, \emph{i.e.}, makeup generation and makeup attack.

As for makeup generation, we first fit the 3D face shape and camera projection matrix for both $I_{src}$ and $I_{ref}$, which enables us to extract the texture maps in the UV space (subsection~\ref{subsection:3d}). Then, a UV generator $G$ is adopted to transfer the texture containing the makeup details of $I_{ref}$ to the texture of the source image $I_{src}$ (subsection~\ref{subsection:texG}). The generated texture with reference makeup styles can be denoted as $T_{t}=G(T_{src},T_{ref})$, where $T_{src}$ and $T_{ref}$ are the UV textures of $I_{src}$ and $I_{ref}$, respectively. After obtaining $T_{t}$, the corresponding generated image $I_{t}$ can be rendered based the source shape $S_{src}$ using a differentiable renderer $R$, i.e., $I_{t}=R(S_{src}, T_{t})$.

As for makeup attack, it is expected to make the generated image $I_{t}$ with cosmetics adversarial. Here, we consider targeted adversarial attack (i.e., impersonation) which aims to improve the identity similarity between $I_{t}$ and the image with the target identity $I_0$. In this way, the generated images are not arbitrarily misclassified as other identities and can be predicted as a specified target identity in an authorized target set. Moreover, the protection of identity information can be easier to achieve due to identity constraints during training. To further improve the transferability of the attack under the black-box condition, we introduce various FR models during training to learn more generalizable adversarial examples with makeup styles in an ensemble manner. 

\begin{figure*}
	\centering
		\includegraphics[width=1\textwidth]{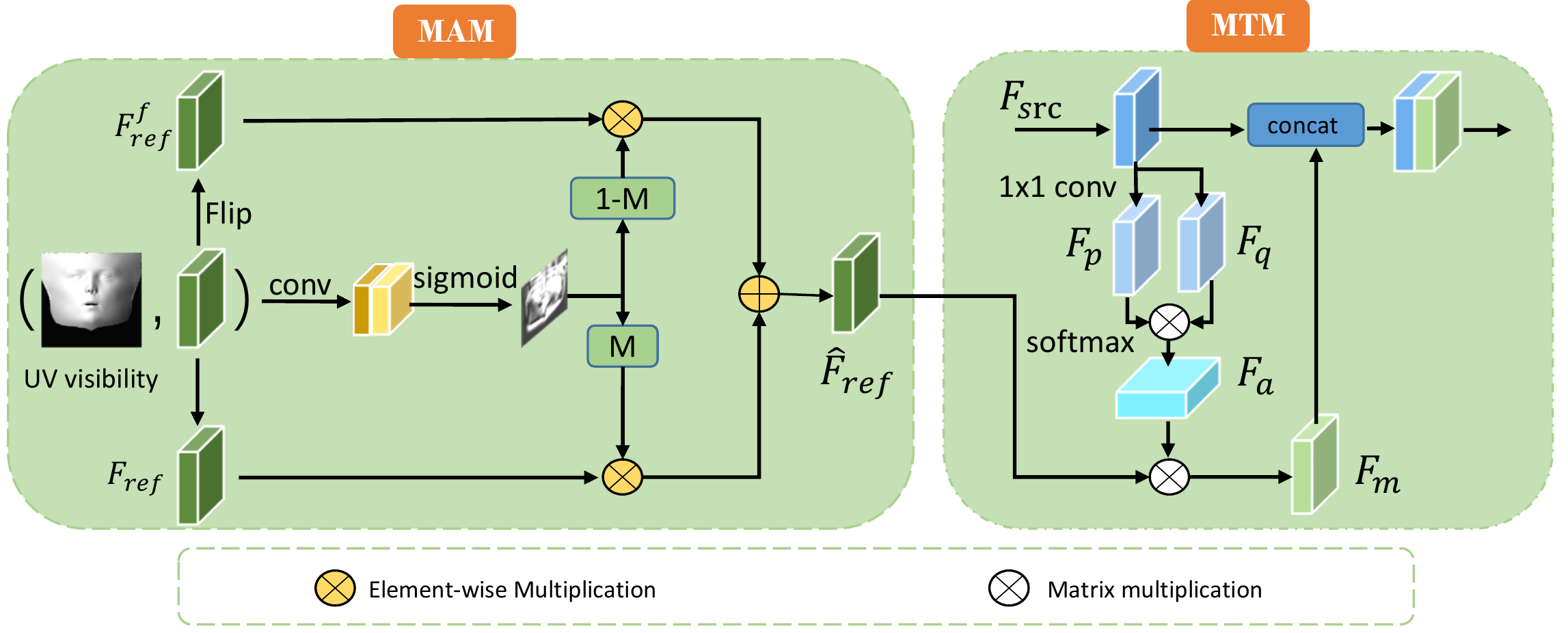}
		\caption{Detailed illustration of the proposed Makeup Adjustment Module (MAM) and Makeup Transfer Module (MTM). MAM is used to adaptively repair unclean reference makeup with the help of a UV visibility map by applying flipping operation and attention fusion, and MTM is responsible for accurate makeup generation.}
	\label{fig:mammtm}
\end{figure*}

\subsection{3D Face Fitting}
\label{subsection:3d}
Blanz and Vetter~\cite{blanz2003face} propose the 3D Morphable Model (3DMM) to describe 3D faces with PCAs.
Here, we estimate the 3DMM coefficients, similar to 3DDFA~\cite{zhu2016face}. We can construct a 3D face shape $\mathbf{S}$ and texture $\mathbf{T}$ for the input faces as:

\begin{equation}
  \begin{aligned}
  \mathbf{S} &=\overline{\mathbf{S}}+\mathbf{A}_{id} \alpha_{id}+\mathbf{A}_{\exp} \alpha_{\exp } \text{,}\\
  \mathbf{T} &=\overline{\mathbf{T}}+\mathbf{A}_{tex} \alpha_{tex} \text{,}
  \end{aligned} 
\end{equation}
where $\mathbf{S} \in \mathbb{R}^{3Q}$ is a 3D face with $Q$ vertices and $\mathbf{T} \in \mathbb{R}^{3Q}$ describes the colors of the corresponding $Q$ vertices. 
Specifically, $\mathbf{S}$ is computed as the linear combination of $\overline{\mathbf{S}}$, $\mathbf{A}_{id}$, and $\mathbf{A}_{\exp}$, where $\overline{\mathbf{S}}$ is the mean face shape, $\mathbf{A}_{id}$ and $\mathbf{A}_{\exp}$ are the PCA bases of identity and expression, respectively. 
Similarly, $\mathbf{T}$ can be obtained by the linear combination of $\overline{\mathbf{T}}$ and $\mathbf{A}_{tex}$, where $\overline{\mathbf{T}}$ is the mean face texture, $\mathbf{A}_{tex}$ is the PCA bases of texture, respectively. 
In this way, any given face could be uniquely described by the coefficients $\alpha_{id}$, $\alpha_{exp}$, $\alpha_{tex}$.

In particular, the texture $\mathbf{T}$ contains the texture intensity value of each vertex $\mathbf{v}_S = (x, y, z)$ in the 3D face mesh. 
In order to obtain the UV texture representation, we project each 3D vertex $\mathbf{v}_S$ onto the UV space using cylindrical unwrapping. 
Suppose that the face mesh has the top pointing up the $y$ axis, the UV space $\mathbf{v}_T = (u, v)$ can be computed as:
\begin{equation}
	v \rightarrow \alpha_{1} \cdot \arctan \left(\frac{x}{z}\right)+\beta_{1}\text{,} \quad u \rightarrow \alpha_{2} \cdot y+\beta_{2} \text{,}
\end{equation}
where $\alpha_{1}, \alpha_{2}, \beta_{1}, \beta_{2}$ are constant scaling and translation parameters to place the unwrapped face into the boundaries of the image. 
Moreover, once we have the estimated 3D shape $\mathbf{S}$ of the facial image, its visible vertices could be computed by z-buffer, which could be utilized to generate a visibility map in the UV space. 

After obtaining the 3D face, we can project it onto the image plane with the weak perspective projection model by: 
\begin{equation}
	V\left(\mathbf{p}_{2d}\right)=f * \mathbf{P_{r}} * \mathbf{R} *\left(\overline{\mathbf{S}}+\mathbf{A}_{id} \boldsymbol{\alpha}_{i d}+\mathbf{A}_{\exp} \boldsymbol{\alpha}_{exp}\right)+\mathbf{t}_{2d} \text{,}
\end{equation}
where $V\left(\mathbf{p}_{2d}\right)$ is the transformation function outputting the 2D positions of 3D vertices. $f$ is the scale factor, $\mathbf{P_r}$ is the orthographic projection matrix $\left(\begin{array}{lll}1 & 0 & 0 \\ 0 & 1 & 0\end{array}\right)$, 
$\mathbf{R}$ is the rotation matrix constructed from Euler angles \emph{pitch, yaw, roll} and $\mathbf{t}_{2d}$ is the translation vector. 
The collection of 3D geometry parameters is $p_{3d}=[f, \emph{pitch, yaw, roll}, \mathbf{t}_{2d}, \alpha_{id}, \alpha_{exp}]$. 
In this paper, we also use the computed Euler angles \emph{yaw} from the rotation matrix $R$ to guide to obtain more accurate and robust makeup ground truths.

\subsection{Makeup Generation}
\label{subsection:texG}
To generate makeup in $I_{src}$, the proposed generator aims to transfer the adversarial makeup in the UV texture space rather than in the original 2D image space. 
This is because the shadow and occlusion in reference makeup images are difficult to deal with in the 2D space, but could be better handled in the UV space by leveraging the symmetric property of human faces. 
Moreover, since the pose and expression variations are implicitly normalized in the UV space, our method can naturally perform more robust and realistic makeup generation than 2D-based methods. 

Specifically, we utilize an encoder-decoder architecture as the backbone of the UV generator. 
Firstly, the encoder $E_{src}$ applies two convolutional layers to distill the texture features $F_{src}$ from the source textures $T_{src}$. 
Then, the makeup features $F_{ref}$ are transferred to the source features $F_{src}$ in the bottleneck part. Finally, by sending the combined features to the decoder, the transferred texture $T_{t}$ is produced. To repair the shadowed or occluded reference makeup and obtain cleaner makeup representation, a reference encoder $E_{ref}$ and two attentive modules, Makeup Adjustment Module (MAM) and Makeup Transfer Module (MTM), are introduced. In particular, the reference encoder $E_{ref}$ has the same structure as the encoder $E_{src}$, but they do not share parameters. Two attentive modules are designed to extract clean makeup features from the reference texture $T_{ref}$ and perform robust adversarial makeup generation.
\begin{figure*}
	\centering
		\includegraphics[width=1\textwidth]{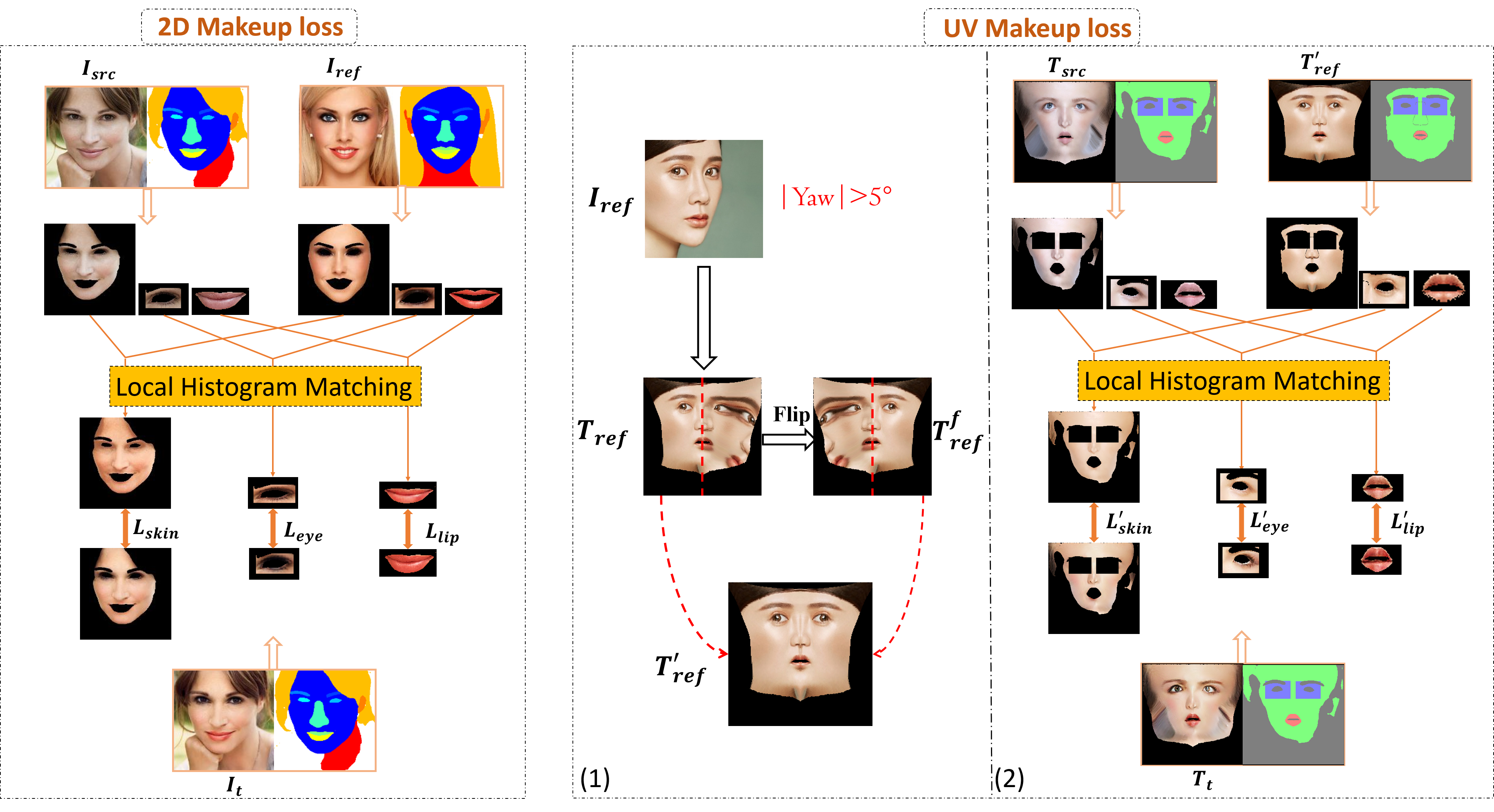}
		\caption{Detailed illustration of the 2D makeup loss and the UV makeup loss. The 2D makeup loss applies the histogram matching in the 2D image space, while the UV makeup loss is used in the UV space. The UV makeup loss can produce more robust makeup supervision, under the condition that non-frontal reference facial images contained incomplete makeup are sampled. For example, when the Euler angles $|\emph{yaw}| \textgreater 5$ degree, there are obvious unclean makeup regions in the reference texture $T_{ref}$. By leveraging the symmetry characteristics in the UV space, we can extract cleaner makeup texture $T_{ref}^{\prime}$ as the makeup supervision from both original and flipped makeup texture, $T_{ref}$ and $T^{f}_{ref}$.}
	\label{fig:allmakeuploss}
\end{figure*}

{\bf Makeup Adjustment Module (MAM).}
The makeup quality of the sampled reference image influences the robustness of the adversarial makeup generation.
For example, when there is shadow and occlusion appearing on the reference image, \emph{e.g.}, the hand covering the face or the self-occlusion caused by facial pose, the extracted reference texture $F_{ref}$ will be contaminated and thus not accurate.
Makeup Adjustment Module (MAM) is designed to adaptively adjust the contaminated makeup regions in feature maps of the reference texture $T_{ref}$ and improve the visual quality of makeup generation. 

As shown in Fig.~\ref{fig:mammtm}, the input consists of the extracted reference features and a UV visibility map of $T_{ref}$. Given that the UV visibility map stores the $z-$coordinate values of the vertex normals, the self-occluded (unclean) regions in the reference image could be detected. 
With the help of the UV visibility map as a prior, MAM produces an attention map $M_{a}$ through several convolutions and a sigmoid operation, which has the same width and height as $F_{ref}$. The attention map $M_{a}$ learns the makeup quality of different regions and gives higher confidence to good quality regions and lower confidence to lower quality regions. For example, the facial regions which are occluded by the hand may have lower attentive weights than other regions.
 
Furthermore, inspired by \cite{Wu_2020_CVPR}, we consider the symmetry of the UV texture of faces and apply a flipping operation to help repair the low-confidence areas. By doing this, we are assuming that facial makeup has a horizontal-symmetry structure, while contaminations are at random and not symmetrical, which is a very frequently true case in the real world.
Therefore, the repaired reference feature $\hat{F}_{ref}$ is obtained by, 

\begin{equation}
  \hat{F}_{ref}=M_{a} \otimes F_{ref} \oplus \left(1-M_{a}\right) \otimes F_{ref}^{f} \text{,}
  \label{con:MAM}
\end{equation}
where $\otimes$ denotes element-wise multiplication, $\oplus$ denotes element-wise addition.
The values of $M_{a}$ are in [0,1]. As shown in Fig.~\ref{fig:visual}, the attentive weights on the contaminated regions are close to zero to eliminate the unreliable textures, which demonstrates the effectiveness of the proposed Makeup Adjustment Module (MAM). 

{\bf Makeup Transfer Module (MTM).}
The refined reference features $\hat{F}_{ref}$ are imported into the Makeup Transfer Module (MTM). 
MTM is a spatial attention module, which aims to transfer the corresponding face makeup of the reference image to the source image precisely.
The proposed structure of MTM is shown in Figure~\ref{fig:mammtm}.
Thanks to the well-alignment of extracted source and reference features in the UV space, each position flowing the network has the similar semantic meaning. Therefore, we directly propose an attention map from the source features to ask for the makeup information of the reference features. 
We calculate the attention map from the source feature by $F_{a}= softmax(F_{p}F_{q})$, where $F_{p}$ and $F_{q}$ are features obtained by two $1 \times 1$ convolution kernels. 
To extract the makeup feature of the corresponding position, the reference feature is embedded by multiplying the attention map, denoted as $F_{m}=F_{a} \otimes \hat{F}_{ref}$. Finally, the concatenation of the source feature and the reference texture is flowed to the decoder for outputting the transfer results. 
In summary, such a design can effectively transfer the makeup from the reference feature into the source feature, by leveraging the spatial-alignment of the UV space and learning a pixel-wise attention map for the reference feature. 

\subsection{Makeup Attack}
\label{subsection:att}
We train our GANs with a makeup attack mechanism to make the generated image $I_t$ with different makeup styles adversarial. 
We conduct targeted adversarial attacks which encourage $I_t$ to have the same visual identity as $I_s$ and recognized as the target identity $I_0$ by unauthorized FR models. 
To this end, we restrict the distance between the target ID features and the ID features of the generated image $I_{t}$ to be as close as possible. 
The distance function $D(\cdot,\cdot)$ measured by the cosine similarity can be expressed as
\begin{equation}
	\begin{aligned}
	D(I_0, I_t) &=\cos \left[H(I_0), H(I_t)\right]\text{,}
	\end{aligned}
\end{equation}
where $H(\cdot)$ is a pre-trained face recognition model.

In order to improve the transferability of adversarial examples under the black-box condition, we implement ensemble training strategy~\cite{liu2016delving,dong2018boosting,yin2021adv,hu2022protecting} to find the common and informative gradient signal with various FR models. Specifically, we choose K pre-trained face recognition models, such as IR152~\cite{he2016deep}, IRSE50~\cite{hu2018squeeze} and FaceNet~\cite{schroff2015facenet}, to serve as the training models, which are proved to have high recognition accuracy on multiple large-scale face datasets. The loss of targeted attack $\mathcal{L}_{att}$ can be described as 
\begin{equation}
	\begin{aligned}
	\mathcal{L}_{att} &=\frac{1}{K}\sum_{k=1}^{K}(1-D_k(I_0, I_t))\text{,}
	\end{aligned}
\end{equation} 
where $D_k$ represents the cosine distance of the $k$-th pre-trained face recognition model. 

\subsection{Loss Functions}
\label{subsection:loss}
In this section, we introduce other loss functions other than the targeted attack loss $\mathcal{L}_{att}$.

{\bf 2D Makeup loss.}
To add supervision for the makeup similarity, we apply the 2D makeup loss proposed by \cite{li2018beautygan} to the rendered face image, which is illustrated in Figure~\ref{fig:allmakeuploss}. The 2D makeup loss computes local histogram matching on three local facial regions, \emph{i.e.}, lips, eye shadow, and skin.
We denote histogram matching between $I_{t}$ and $I_{ref}$ as $HM(I_{t} \otimes M_{item},I_{ref} \otimes M_{item})$, where $M_{item}$ is the local region obtained by a face parsing model, $item \in \{lips, eye shadow, skin\}$. 
As a kind of ground truth, $HM(I_{t} \otimes M_{item},I_{ref} \otimes M_{item})$ restricts the rendered image and the reference image to have similar makeup style in the locations of $M_{item}$.

The local histogram matching loss is expressed as 
\begin{equation}
	\begin{aligned}
		\hspace{-4mm}\mathcal{L}_{item}\!=\!\left\|HM(I_{t}\!\otimes\!M_{item},I_{ref}\!\otimes\!M_{item})\!-I_{t}\!\otimes\!M_{item}\right\|_{2} \text{.}
\end{aligned}
\end{equation}

The overall 2D makeup loss is expressed as
\begin{equation}
	\begin{aligned}
\mathcal{L}_{makeup}=\lambda_{1}\mathcal{L}_{lips}+\lambda_{2}\mathcal{L}_{eye}+\lambda_{3}\mathcal{L}_{skin} \text{.}
	\end{aligned}
	\label{con:makeuploss}
\end{equation}

{\bf UV Makeup Loss.}
The above 2D makeup loss adopted to the rendered images has one limitation. It may not provide accurate ground truth when non-frontal reference facial images contained incomplete makeup are sampled. To address this problem, our 3DAM-GAN benefits from the symmetry of UV space and applies a novel UV makeup loss in the UV space. 
We first compute Euler angles \emph{yaw} from the rotation matrix $R$, described in subsection~\ref{subsection:3d}. 
To some extent, when the value of \emph{yaw} is not in [-5,5] degrees, there are obvious incomplete/unclean makeup regions due to self-occlusion, as shown in Figure~\ref{fig:allmakeuploss}. In this case, we extract accurate makeup information from both original and flipped makeup textures to obtain more accurate makeup supervision. 

The new UV local histogram loss can be formulated as 
\begin{equation}
	\begin{aligned}
		\hspace{-3.5mm}\mathcal{L}^{\prime}_{item}\!=\!\left\|HM(T_{t}\!\otimes\!M^{\prime}_{item},T^{\prime}_{ref}\!\otimes\!M^{\prime}_{item})\!-\!T_{t}\!\otimes\!M^{\prime}_{item}\right\|_{2} \text{.}
	\end{aligned}
\end{equation}
where $T^{\prime}_{ref}$ is the repaired reference makeup texture and $T_{t}$ is the generated adversarial makeup texture. $M^{\prime}_{item}$ is the corresponding UV region mask, $item \in \{lips, eye shadow, skin\}$. 

The overall UV makeup loss is formulated as
\begin{equation}
	\begin{aligned}
\mathcal{L}^{uv}_{makeup}=\lambda_{1}\mathcal{L}^{\prime}_{lips}+\lambda_{2}\mathcal{L}^{\prime}_{eye}+\lambda_{3}\mathcal{L}^{\prime}_{skin} \text{,}
	\end{aligned}
	\label{con:sysmakeuploss}
\end{equation}

{\bf Perceptual loss.}
Perceptual loss~\cite{johnson2016perceptual} is defined by the distance of the high-dimensional features between two images. 
In order to preserve the facial identity and improve the visual quality of the transferred image, we adopt it to measure the distance between the generated texture $T_{t}$ and the source texture $T_{src}$ in the feature space. 
Here, a pre-trained $VGG$-16 model~\cite{vgg} on ImageNet is used as the feature extractor. 
Let $F^{per}_i$ represent the output of $i^{th}$ layer of the feature extractor.
The perceptual loss can be expressed as follows: 

\begin{equation}
	\begin{aligned}
\mathcal{L}_{per}&=\left\|F^{per}_i(T_t)-F^{per}_i(T_{src})\right\|_{2}\text{,}\\
	\end{aligned}
\end{equation}
where $\|\cdot\|_{2}$ represents the L2-Norm.

\begin{algorithm}[t]
	\small
	\caption{The training process of the proposed 3DAM-GAN}
	\label{alg:attack}
	\begin{algorithmic}[1]
		\Require Source image $I_s \in D_s$; reference image $I_r \in D_r$; target image $I_0$; pre-trained FR model bank $\mathcal{H}=\{{H_k}\}^{K}_{k=1}$ ; generator $G$; discriminators $D^S_{tex}$, $D^R_{tex}$, $D_{img}$; iteration times $T$.
		\Ensure
		Model parameters $\theta^{*}_G$, $\theta^{*}_{D^S_{tex}}$, $\theta^{*}_{D^R_{tex}}$, $\theta^{*}_{D_{img}}$.
		\For {$t = 0$; $t\textless T$; $t\leftarrow t+1$}
		\State Sample source image $I_s \sim \{D_s\}$
		\State Sample reference image $I_r \sim \{D_r\}$
		\State Update $D^S_{tex}$, $D^R_{tex}$, $D_{img}$ w.r.t $\mathcal{L_D}$ in Eq. (\ref{con:totalloss}): \\
		$\qquad \qquad \theta_{D^S_{tex}}^{\prime} \leftarrow \theta_{D^S_{tex}}-\nabla \theta_{D^S_{tex}} \mathcal{L}_{D}\left(\theta_{D^S_{tex}}\right)$\\
		$\qquad \qquad \theta_{D^R_{tex}}^{\prime} \leftarrow \theta_{D^R_{tex}}-\nabla \theta_{D^R_{tex}} \mathcal{L}_{D}\left(\theta_{D^R_{tex}}\right)$\\
		$\qquad \qquad \theta_{D_{img}}^{\prime} \leftarrow \theta_{D_{img}}-\nabla \theta_{D_{img}} \mathcal{L}_{D}\left(\theta_{D{img}}\right)$
		\State Update $G$ with respect to $\mathcal{L_G}$ in Eq. (\ref{con:totalloss}):\\
		$\qquad \qquad \theta_{G}^{\prime} \leftarrow \theta_{G}-\nabla \theta_{G} \mathcal{L}_{G}\left(\theta_{G}\right)$
		\EndFor \\
		\Return $G$ after training of $T$ iterations.
	\end{algorithmic}
\end{algorithm}
\begin{table*}[t]
	\caption{Evaluation of black-box attack success rate (ASR) at 0.01 FAR/0.001 FAR for the LADN and CelebA-HQ dataset. For each column, we consider the written model as the target victim model and the remaining three models are used for ensemble training. Our approach achieves high ASR results on all four target models and two different FARs, yielding the high black-box transferability and robustness.}
	\centering
	\resizebox{1\textwidth}{!}{
	\begin{tabular}{c|cccc|cccc}
	\hline & \multicolumn{4}{c|}{ LADN Dataset } & \multicolumn{4}{c}{ CelebA-HQ Dataset } \\
	\hline & IRSE50 & IR152 & FaceNet & MobileFace & IRSE50 & IR152 & FaceNet & MobileFace \\
	\hline Clean & \cellcolor{red!5}$26.59/12.16$ & \cellcolor{red!5}$15.15/6.28$ & \cellcolor{red!5}$9.46/1.80$ & \cellcolor{red!5}$25.63/8.14$ & \cellcolor{red!5}$6.10/1.04$ & \cellcolor{red!5}$3.68/0.76$ & \cellcolor{red!5}$2.44/0.08$ & \cellcolor{red!5}$5.88/0.94$ \\
	\hline PGD~\cite{kurakin2017adversarial} & $61.02/38.68$ & $36.47/21.14$ & $18.20/3.65$ & $55.21/29.04$ & $30.08/12.18$ & $16.30/5.94$ & $7.08/0.38$ & $25.72/7.86$ \\
	FGSM~\cite{goodfellow2014explaining} & $63.89/42.22$ & $38.68/22.81$ & $20.36/4.07$ & $55.69/29.46$ & $35.72/16.06$ & $19.14/7.34$ & $8.08/0.52$ & $28.68/9.10$ \\
	MI-FGSM~\cite{dong2018boosting} & $72.69/51.44$ & $42.10/26.59$ & $22.93/5.21$ & $63.11/37.25$ & $42.22/21.28$ & $21.58/8.78$ & $8.98/0.66$ & $34.58/12.10$ \\
	TI-DIM~\cite{dong2019evading} & $75.45/53.53$ & \cellcolor{red!15}$42.93/27.25$ & \cellcolor{red!15}$28.32/7.66$ & $66.71/39.94$ & $45.08/23.20$ & \cellcolor{red!15}$22.60/9.64$ & \cellcolor{red!15}$13.48/1.10$ & $36.14/13.12$ \\
	TIP-IM~\cite{yang2021towards} & $34.13/17.07$ & $38.38/23.89$ & $12.04/2.34$ & $29.28/10.18$ & $9.30/2.46$ & $16.22/6.02$ & $3.78/0.12$ & $7.40/1.16$ \\
	Adv-Makeup~\cite{yin2021adv} & \cellcolor{red!15}$87.13/67.49$ & $34.07/18.56$ & $26.41/4.37$ & \cellcolor{red!15}$87.31/71.91$ & \cellcolor{red!15}$65.60/40.50$ & $15.42/5.62$ & $11.68/0.92$ & \cellcolor{red!35}$70.15/43.84$ \\
	AMT-GAN~\cite{hu2022protecting} & \cellcolor{red!25}$96.47/89.28$ & \cellcolor{red!25}$65.93/47.12$ & \cellcolor{red!25}$72.51/25.45$ & \cellcolor{red!35}$88.98/64.01$ & \cellcolor{red!25}$77.99/54.20$ & \cellcolor{red!25}$36.27/16.77$ & \cellcolor{red!25}$47.10/8.49$ & \cellcolor{red!15}$47.22/16.11$ \\
	\hline Ours & \cellcolor{red!35}$98.38/91.83$ & \cellcolor{red!35}$71.47/55.32$ & \cellcolor{red!35}$72.98/26.07$ & \cellcolor{red!25}$88.65/61.62$ & \cellcolor{red!35}$89.44/69.63$ & \cellcolor{red!35}$57.04/35.56$ & \cellcolor{red!35}$64.33/16.30$ & \cellcolor{red!25}$57.24/22.90$ \\
	\hline
	\end{tabular}}
	\label{tab:table1}
\end{table*}

{\bf Adversarial loss.}
To further improve the fidelity of the generated results, we introduce adversarial losses during the training. 
We utilize two domain discriminators $D_{tex}^{S}$, $D_{tex}^{R}$ for the generated UV texture maps and one real-fake discriminator $D_{img}$ for the rendered face. 
Specifically, $D_{tex}^{S}$ aims to distinguish the generated texture from the source texture domain $S$, $D_{tex}^{R}$ aims to distinguish the generated texture from the reference texture domain $R$, and $D_{img}$ is trained to tell whether the generated outputs are real or fake.

Mathematically, the adversarial losses, $\mathcal{L}_{adv}^{D_{tex}}$ and $\mathcal{L}_{adv}^{G_{tex}}$ are formulated as: 
\begin{equation}
	\begin{aligned}
  \mathcal{L}_{adv}^{D_{tex}}=&-\mathbb{E}\left[\log D_{tex}^{S}(T_{src})\right]-\mathbb{E}\left[\log D_{tex}^{R}(T_{ref})\right] \\
  &-\mathbb{E}\left[\log \left(1-D_{tex}^{S}(G(T_{ref}, T_{src}))\right)\right] \\
  &-\mathbb{E}\left[\log \left(1-D_{tex}^{R}(G(T_{src}, T_{ref}))\right)\right] \text{,}\\
  \mathcal{L}_{adv}^{G_{tex}}=&-\mathbb{E}\left[\log \left(D_{tex}^{S}(G(T_{ref}, T_{src}))\right)\right] \\
  &-\mathbb{E}\left[\log \left(D_{tex}^{R}(G(T_{src}, T_{ref}))\right)\right] \text{.}
  \end{aligned}
\end{equation}

The adversarial losses $\mathcal{L}_{adv}^{D_{img}}$ and $\mathcal{L}_{adv}^{G_{img}}$ are formulated as: 
\begin{equation}
	\begin{aligned}
  \mathcal{L}_{adv}^{D_{img}}=&-\mathbb{E}\left[\log D_{img}(I_{src})\right]-\mathbb{E}\left[\log D_{img}(I_{ref})\right] \\
  &-\mathbb{E}\left[\log \left(1-D_{img}(R(S_{ref},G(I_{ref}, I_{src})))\right)\right] \\
  &-\mathbb{E}\left[\log \left(1-D_{img}(R(S_{src},G(I_{src}, I_{ref})))\right)\right] \text{,}\\
  \mathcal{L}_{adv}^{G_{img}}=&-\mathbb{E}\left[\log \left(D_{img}(R(S_{ref},G(I_{ref}, I_{src})))\right)\right] \\
  &-\mathbb{E}\left[\log \left(D_{img}(R(S_{src},G(I_{src}, I_{ref})))\right)\right]\text{.}
  \end{aligned}
\end{equation}

{\bf Total Loss Function.}
By combining all the above defined losses, the total loss functions can be expressed as follows:
\begin{equation}
  \begin{aligned}
  \mathcal{L}_{G} &= \lambda_{a}\mathcal{L}_{adv}^{G_{tex}} + \lambda_{a}\mathcal{L}_{adv}^{G_{img}} + \lambda_{m}\mathcal{L}_{makeup} \\
&+ \lambda^s_{m}\mathcal{L}^s_{makeup} + \lambda_{p}\mathcal{L}_{per} + \lambda_{t}\mathcal{L}_{att}   \text{,}\\
  \mathcal{L}_{D} &= \lambda_{a}\mathcal{L}_{adv}^{D_{tex}} + \lambda_{a}\mathcal{L}_{adv}^{D_{img}} \text{,}
  \end{aligned}
  \label{con:totalloss}
\end{equation}
where $\mathcal{L}_{G}$ is the loss for training the UV texture generator. $\mathcal{L}_{D}$ is the loss for training the discriminators. 
$\lambda_a$, $\lambda_m$, $\lambda^s_m$, $\lambda_p$, $\lambda_t$ are the trade-off parameters to balance the different loss terms. The entire training process is illustrated in Alg.\ref{alg:attack}.

\section{Experiments}
\label{sec:exp}
\subsection{Experimental Settings}
{\bf Datasets.}
In our experiments, we use the Makeup Transfer (MT) dataset as the training dataset, which is currently the largest facial makeup dataset proposed by \cite{li2018beautygan}. 
It consists of 1,115 non-makeup images, and 2,719 makeup images with diverse makeup styles from subtle to heavy, including Korean makeup style, flashy makeup style, and smoky-eyes makeup style. 
To evaluate the effectiveness and robustness of our method, we use two additional public face datasets as the test sets: 
(1) LADN~\cite{gu2019ladn}, a high-quality makeup dataset that consists of 334 non-makeup images and 302 makeup images. Following \cite{hu2022protecting}, we use 334 non-makeup images as the test images. 
(2) CelebA-HQ~\cite{karras2018progressive}, a large-scale face image dataset that has 30,000 high-resolution face images. We randomly choose 1,000 different face images as the test identities. 
For both of the test sets, we randomly choose 5 target images for evaluation and report the average results across all identity pairs. Note that the chosen target images are only used during training to provide identity constraints and guidance, making identity protection for source images easier to achieve. In practical testing, the model trained with a single target image $I_0$, can apply adversarial makeup to any input source image $I_{src}$ to generate a protected image $I_t$.

{\bf Benchmark Methods.}
We conduct comparisons with different benchmark schemes of adversarial attacks, \emph{i.e.}, PGD~\cite{kurakin2017adversarial}, FGSM~\cite{goodfellow2014explaining}, MI-FGSM~\cite{dong2018boosting}, TI-DIM~\cite{dong2019evading}, TIP-IM~\cite{yang2021towards}, Adv-Makeup~\cite{yin2021adv} and AMT-GAN~\cite{hu2022protecting}. Specifically, PGD~\cite{kurakin2017adversarial}, FGSM~\cite{goodfellow2014explaining}, MI-FGSM~\cite{dong2018boosting}, TI-DIM~\cite{dong2019evading} and TIP-IM~\cite{yang2021towards} are popular gradient-based methods with strong attack ability, where TIP-IM~\cite{yang2021towards} is specifically proposed for facial privacy protection. Adv-Makeup~\cite{yin2021adv} and AMT-GAN~\cite{hu2022protecting} are recent works that also derive adversarial facial images with makeup styles. 
For fair comparisons, we follow the official experimental settings of all compared methods.
\begin{table}[t!]
	\centering
	\caption{Quantitative comparison results in terms of FID, SSIM and PSNR.}
	\begin{tabular}{c|ccccc}
		\toprule  Methods& FID~($\downarrow$) & SSIM~($\uparrow$) & PSNR~($\uparrow$) \\
		\midrule  Adv-Makeup~\cite{yin2021adv} &$3.05$ & $0.98$ & $33.76$\\
				  AMT-GAN~\cite{hu2022protecting} & $28.22$ & $0.82$ & $19.59$\\
		\midrule  PSGAN~\cite{jiang2020psgan} & $22.19$ & $0.85$ & $19.77$\\
				  PSGAN++~\cite{liu2021psgan++} & $23.07$ & $0.84$ & $19.02$\\
				  LADN~\cite{gu2019ladn} & $73.08$ & $0.65$ & $14.65$\\
				  SOGAN~\cite{lyu2021sogan} & $32.88$ & $0.91$ & $20.88$\\
		\midrule  Ours & $\bf{26.78}$ & $\bf{0.91}$ & $\bf{20.82}$\\
		  		  w/o MAM & $31.52$ & $0.90$ & $19.63$\\
		  		  w/o UV makeup loss & $35.63$ & $0.89$ & $19.29$\\
		\bottomrule
		\end{tabular}
\label{tab:table2}
\end{table}

\begin{figure*}[tp]
	\centering
		\includegraphics[width=1\linewidth]{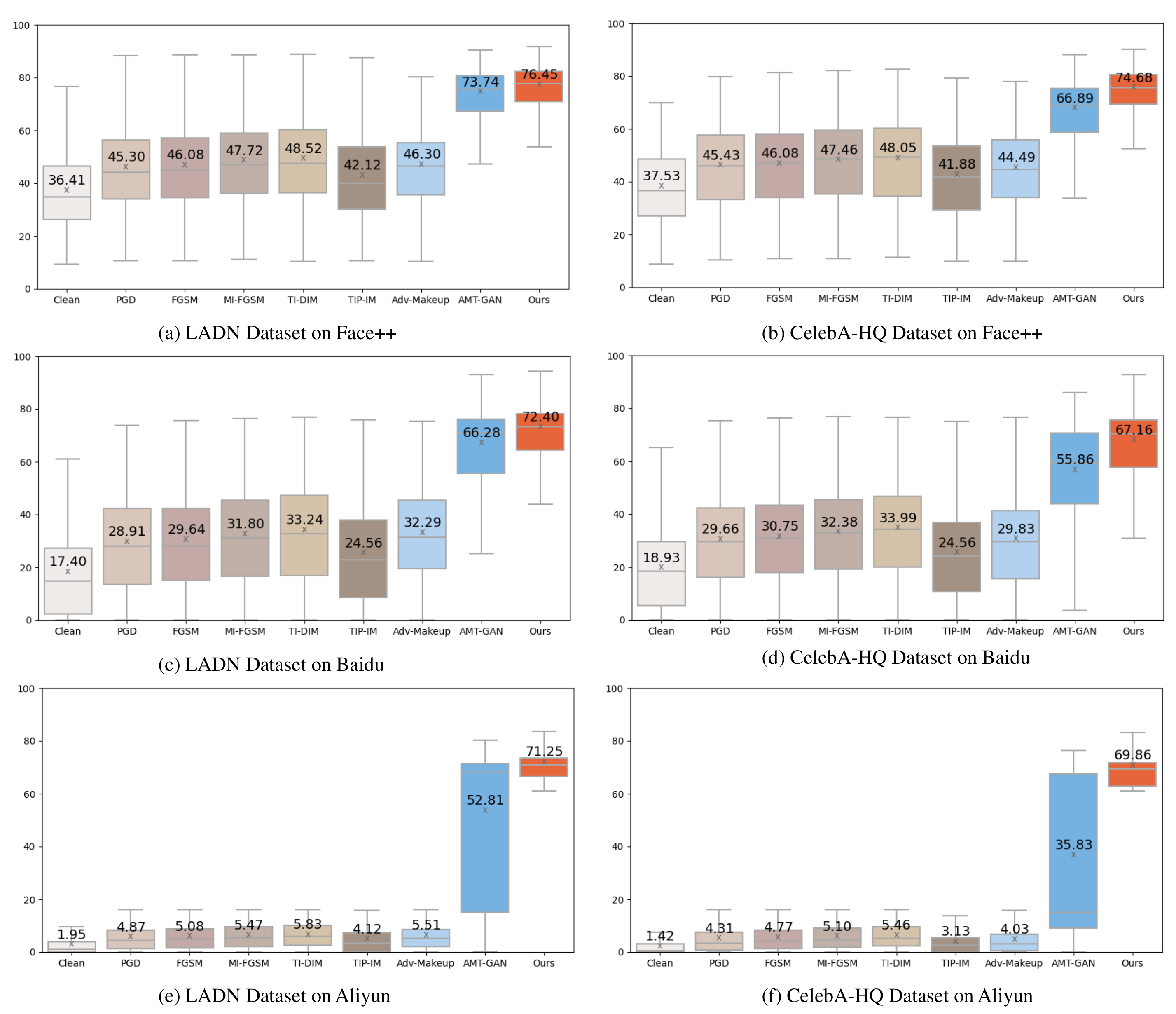}
		\caption{Comparison of confidence scores returned from commercial FR APIs over the LADN dataset and CelebA-HQ dataset. All other methods suffer a significant degradation in Aliyun, but our approach maintains high average confidence scores for all APIs.}
	\label{fig:faceapi}
\end{figure*}

{\bf Target Models.}
We conduct experiments on four public FR models, including IRSE50~\cite{hu2018squeeze}, IR152~\cite{he2016deep}, FaceNet~\cite{schroff2015facenet} and MobileFace~\cite{deng2019arcface}. The pre-trained models for IRSE50 and MobileFace are obtained from the InsightFace$\_$Pytorch library\footnote[5]{\url{https://github.com/TreB1eN/InsightFace_Pytorch}}, for IR152 from the face.evlLVe library\footnote[6]{\url{https://github.com/ZhaoJ9014/face.evoLVe}}, and for FaceNet from the facenet-pytorch library\footnote[7]{\url{https://github.com/timesler/facenet-pytorch}}. In particular, three of them are chosen for ensemble training and the remaining unseen one is the black-box FR model for transferable testing. Moreover, to further explore the reliability of our method, we evaluate on three commercial state-of-the-art online FR APIs of unknown FR models, which include Face++, Baidu, and Aliyun. It should be emphasized that, in order to verify the transferability and practicality of our method, all the tests are conducted on FR models that are not encountered during training.

{\bf Implementation Details.}
Our model is trained on 1 NVIDIA RTX 2080Ti GPU, with PyTorch3D (v0.2.0) \cite{ravi2020accelerating}. Some preliminary prototyping is conducted on the MindSpore framework during our implementation.
For the optimizer, we choose ADAM~\cite{kingma2014adam} with $\beta_1$ = 0, $\beta_2$ = 0.9. We use a batch size of 1 and a constant learning rate of 0.0002. The trade-off parameters in Eq.~(\ref{con:sysmakeuploss}) are set as $\lambda_1$ = $\lambda_2$ = 1, $\lambda_3$ = 0.1. The trade-off parameters in Eq.~(\ref{con:totalloss}) are set as $\lambda_a$ = $\lambda_m$ = 1, $\lambda_t$ = 10, $\lambda_p$ = 5e-3.

{\bf Evaluation Metrics.}
To evaluate the effectiveness of the privacy protection of different methods, we report the \emph{attack success rate} (ASR), which is the common evaluation metric in prior works of the privacy protection~\cite{yin2021adv,xiao2021improving,hu2022protecting}. It is computed by
\begin{equation}
	ASR=\frac{\sum_{i=1}^{N} 1_{\tau}\left(\cos \left[H\left(I_{t}^{i}\right), H\left({I}_{0}\right)\right]>\tau\right)}{N} \times 100 \%\text{,}
\end{equation}
where H($\cdot$) represents the target FR model, $I_0$ is the chosen target identity, and $I_t^i$ represents the generated $i$-th adversarial example. $1_{\tau}$ is the indicator function of attack success.
For black-box evaluations, the proportion of similarity comparisons above the threshold $\tau$ is considered as the attack success. 
Here, the values of $\tau$ are set as the thresholds at 0.01 and 0.001 FAR (False Acceptance Rate) for all FR models as most face recognition works do, \emph{i.e.}, 0.241 and 0.313 for IRSE50, 0.167 and 0.228 for IR152, 0.302 and 0.381 for MobileFace, 0.409 and 0.591 for FaceNet. 
For commercial FR APIs, the \emph{confidence scores} reported by FR servers are directly used for evaluations, which represents the confidence of matching between the generated images and the target image. Here, the thresholds recommended by the service providers are 69.10~(Face++), 80.00~(Baidu) and 69.00~(Aliyun).

In addition, to evaluate the image generation quality, we leverage \emph{Fréchet Inception Distance} (FID)~\cite{heusel2017gans}, \emph{Peak Signal-to-Noise Ratio} (PSNR) and \emph{Structural Similarity Index Measure} (SSIM)~\cite{wang2004image} as the evaluation metrics. Note that a lower FID value indicates superior performance, while larger PSNR and SSIM values mean better image quality.

\begin{table*}[t]
	\caption{Evaluation of protection success rate (PSR, higher is better) at 0.01 FAR/0.001 FAR for the LFW and CelebA-HQ dataset under face verification setting. For each column, we consider the header model as the target victim model and the remaining three models are used for ensemble training. Our approach achieves the highest PSR results on all four target models and two different FARs, yielding the highest identity protection ability.}
	\centering
	
	\resizebox{1\textwidth}{!}{
	\begin{tabular}{c|cccc|cccc}
	\hline & \multicolumn{4}{c|}{ LFW Dataset } & \multicolumn{4}{c}{ CelebA-HQ Dataset } \\
	\hline & IRSE50 & IR152 & FaceNet & MobileFace & IRSE50 & IR152 & FaceNet & MobileFace \\
	\hline Clean & \cellcolor{red!5}$2.11/3.75$ & \cellcolor{red!5}$1.55/2.19$ & \cellcolor{red!5}$1.62/7.50$ & \cellcolor{red!5}$5.73/13.83$ & \cellcolor{red!5}$6.36/10.81$ & \cellcolor{red!5}$6.40/8.70$ & \cellcolor{red!5}$6.74/19.72$ & \cellcolor{red!5}$11.87/23.76$ \\
	\hline PGD~\cite{kurakin2017adversarial} & $6.07/11.06$ & \cellcolor{red!15}$6.31/9.51$ & $6.16/19.59$ & $12.10/24.45$ & $7.00/12.58$ & $7.34/10.97$ & $7.10/22.06$ & $13.33/26.30$ \\
	FGSM~\cite{goodfellow2014explaining} & $4.17/10.15$ & $3.66/7.53$ & $2.98/16.75$ & $11.38/26.30$ & $10.73/21.06$ & $11.88/19.64$ & $9.69/33.03$ & $18.48/35.16$ \\
	MI-FGSM~\cite{dong2018boosting} & $2.50/5.25$ & $1.96/3.31$ & $1.92/9.79$ & $7.26/17.93$ & $7.29/13.27$ & $7.58/11.54$ & $7.26/22.95$ & $13.59/26.85$ \\
	TI-DIM~\cite{dong2019evading} & $2.61/5.70$ & $2.09/3.38$ & $2.23/12.10$ & $7.40/18.14$ & $7.64/13.99$ & $7.72/11.99$ & $8.03/26.51$ & $13.79/27.17$ \\
	TIP-IM~\cite{yang2021towards} & $3.69/8.37$ & $4.57/8.73$ & $3.78/19.21$ & $8.89/21.53$ & $10.50/19.96$ & \cellcolor{red!15}$14.09/23.04$ & \cellcolor{red!15}$12.32/40.21$ & $17.28/32.77$ \\
	Adv-Makeup~\cite{yin2021adv} & \cellcolor{red!15}$7.29/17.47$ & $3.35/7.13$ & \cellcolor{red!15}$4.94/26.72$ & \cellcolor{red!15}$23.38/45.27$ & \cellcolor{red!15}$13.17/26.27$ & $10.48/17.28$ & $11.16/38.56$ & \cellcolor{red!15}$26.98/48.30$ \\
	AMT-GAN~\cite{hu2022protecting} & \cellcolor{red!25}$18.97/36.87$ & \cellcolor{red!25}$22.40/38.90$ & \cellcolor{red!25}$26.68/67.02$ & \cellcolor{red!25}$31.54/54.51$ & \cellcolor{red!25}$35.34/57.30$ & \cellcolor{red!25}$38.25/56.77$ & \cellcolor{red!25}$43.99/81.91$ & \cellcolor{red!25}$43.20/66.69$ \\
	\hline Ours & \cellcolor{red!35}$56.96/79.00$ & \cellcolor{red!35}$60.96/80.21$ & \cellcolor{red!35}$82.78/98.02$ & \cellcolor{red!35}$66.60/85.86$ & \cellcolor{red!35}$60.42/81.69$ & \cellcolor{red!35}$67.30/84.60$ & \cellcolor{red!35}$82.11/98.40$ & \cellcolor{red!35}$58.67/79.98$ \\
	\hline
	\end{tabular}}
	\label{tab:table_pro}
\end{table*}

\begin{table*}[t]
	\caption{Evaluation of Rank-1 and Rank-5 protection success rate (higher is better) for the LFW and CelebA-HQ dataset under face identification setting. Our approach achieves the highest results on all four target models, yielding the highest identity protection ability.}
	\centering
	\label{tab:table_rank}
	
	\resizebox{1\textwidth}{!}{
	\begin{tabular}{c|cccccccc|cccccccc}
	\hline
	\multicolumn{1}{l|}{} & \multicolumn{8}{c|}{LFW Dataset}                                                                                       & \multicolumn{8}{c}{CelebA-HQ Dataset}                                                                                 \\ \hline
	\multicolumn{1}{l|}{} & \multicolumn{2}{c}{IRSE50} & \multicolumn{2}{c}{IR152} & \multicolumn{2}{c}{FaceNet} & \multicolumn{2}{c|}{MobileFace} & \multicolumn{2}{c}{IRSE50} & \multicolumn{2}{c}{IR152} & \multicolumn{2}{c}{FaceNet} & \multicolumn{2}{c}{MobileFace} \\
	\multicolumn{1}{l|}{}               & Rank-1   & Rank-5  & Rank-1  & Rank-5  & Rank-1   & Rank-5   & Rank-1     & Rank-5     & Rank-1   & Rank-5  & Rank-1  & Rank-5  & Rank-1   & Rank-5   & Rank-1     & Rank-5    \\ \hline
	Clean                               & \cellcolor{red!5}4.73     & \cellcolor{red!5}2.39    & \cellcolor{red!5}2.58    & \cellcolor{red!5}1.39    & \cellcolor{red!5}4.06     & \cellcolor{red!5}1.81     & \cellcolor{red!5}15.02      & \cellcolor{red!5}9.21 & \cellcolor{red!5}17.23 & \cellcolor{red!5}10.93 & \cellcolor{red!5}13.08 & \cellcolor{red!5}8.06 & \cellcolor{red!5}14.84 & \cellcolor{red!5}8.67 & \cellcolor{red!5}36.52 & \cellcolor{red!5}25.70\\ \hline
	PGD~\cite{kurakin2017adversarial}   & 6.03     & 3.02    & 4.05    & 1.90    & 4.86     & 2.07     & 19.29      & 11.51      & 20.99    & 13.25   & 17.07   & 10.74   & 17.03    & 9.91     & 41.94      & 30.49     \\
	FGSM~\cite{goodfellow2014explaining}& 11.44    & 5.81    & 9.56    & 4.40    & 9.14     & 3.76     & 26.63      & 16.38      & 31.59    & 20.82   & 29.39   & 19.02   & 24.76    & 14.63    & 51.54      & 38.44     \\
	MI-FGSM~\cite{dong2018boosting}     & 6.57     & 3.29    & 4.39    & 2.13    & 5.19     & 2.16     & 19.71      & 11.82      & 22.01    & 14.06   & 17.91   & 11.21   & 17.72    & 10.32    & 42.53      & 30.90     \\
	TI-DIM~\cite{dong2019evading}       & 7.12     & 3.51    & 4.69    & 2.23    & 6.50     & 2.53     & 19.82      & 11.75      & 23.05    & 14.76   & 18.80   & 11.77   & 20.24    & 11.87    & 42.64      & 31.12     \\
	TIP-IM~\cite{yang2021towards}       & 9.05     & 4.74    & \cellcolor{red!15}10.96   & \cellcolor{red!15}5.63    & 11.94     & 5.48     & 21.88      & 13.42      & 29.04     & 19.30    & \cellcolor{red!15}35.48   & \cellcolor{red!15}24.43    & \cellcolor{red!15}30.74    & \cellcolor{red!15}19.24     & 46.21      & 34.39     \\
	Adv-makeup~\cite{yin2021adv}        & \cellcolor{red!15}17.04    & \cellcolor{red!15}9.34    & 9.95    & 4.70    & \cellcolor{red!15}15.72    & \cellcolor{red!15}7.15     & \cellcolor{red!15}35.58      & \cellcolor{red!15}22.31      & \cellcolor{red!15}36.84    & \cellcolor{red!15}24.17    & 26.89    & 17.12    & 29.66    & 17.48     & \cellcolor{red!15}61.94      & \cellcolor{red!15}46.28     \\
	AMT-GAN~\cite{hu2022protecting}     & \cellcolor{red!25}35.66    & \cellcolor{red!25}21.91   & \cellcolor{red!25}46.46   & \cellcolor{red!25}30.43   & \cellcolor{red!25}48.84    & \cellcolor{red!25}31.27    & \cellcolor{red!25}48.54      & \cellcolor{red!25}32.83      & \cellcolor{red!25}71.81    & \cellcolor{red!25}58.93   & \cellcolor{red!25}75.28   & \cellcolor{red!25}62.36   & \cellcolor{red!25}76.04    & \cellcolor{red!25}62.59    & \cellcolor{red!25}79.89      & \cellcolor{red!25}68.37     \\ \hline
	Ours                                & \cellcolor{red!35}80.41    & \cellcolor{red!35}65.99   & \cellcolor{red!35}86.17   & \cellcolor{red!35}73.72   & \cellcolor{red!35}93.17    & \cellcolor{red!35}84.92    & \cellcolor{red!35}88.95      & \cellcolor{red!35}77.73      & \cellcolor{red!35}95.52    & \cellcolor{red!35}90.45   & \cellcolor{red!35}96.22   & \cellcolor{red!35}91.41   & \cellcolor{red!35}97.84    & \cellcolor{red!35}94.81    & \cellcolor{red!35}96.61      & \cellcolor{red!35}92.83     \\ \hline
	\end{tabular}}
\end{table*}

\begin{figure*}[tp]
	\centering
		\includegraphics[width=0.9\linewidth]{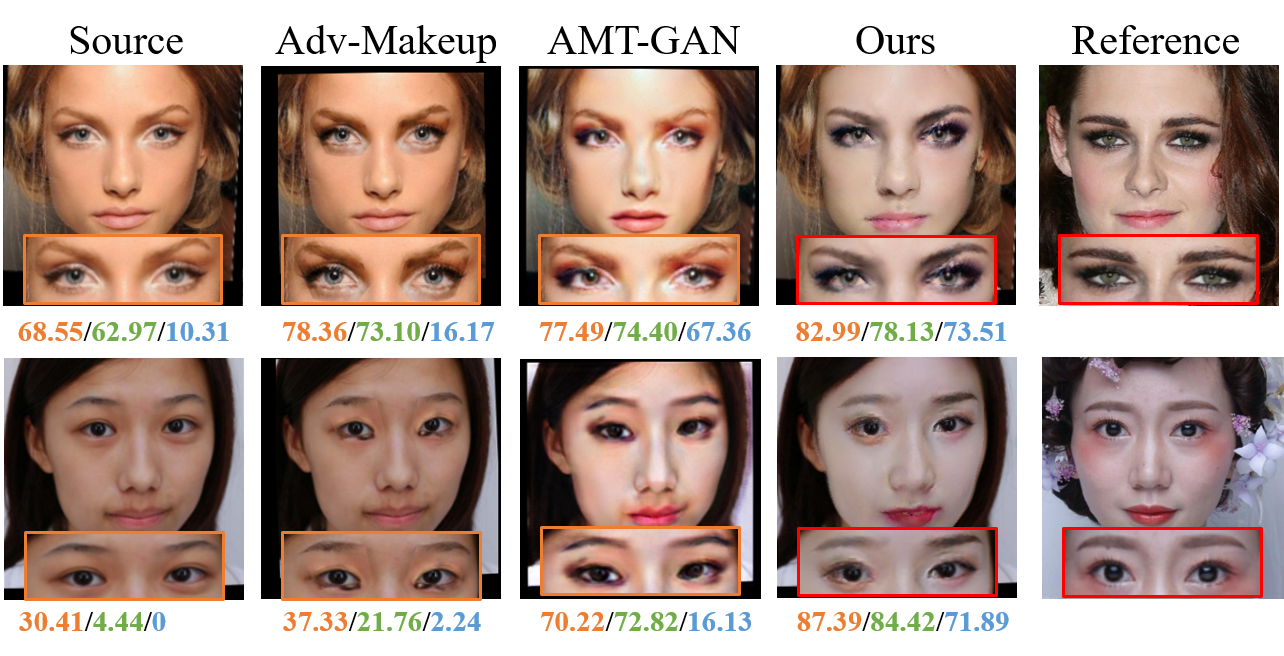}
		\caption{Qualitative comparison on visual quality with Adv-Makeup and AMT-GAN. The three numbers below each face image are the confidence scores returned by the commercial online FR systems, Face++, Baidu and Aliyun.}
	\label{fig:quali}
\end{figure*}

\begin{table*}[tp]
	\centering
	\caption{Comparison of time efficiency of our method with other competitors. We can achieve faster computation time compared with PGD, MI-FGSM, and TI-DIM.}
	\resizebox{1\textwidth}{!}{
	\begin{tabular}{c|cccccccc}
	\toprule Methods & PGD & FGSM & MI-FGSM & TI-DIM & TIP-IM & Adv-Makeup & AMT-GAN & Ours \\
	\midrule Time (sec./image) &2.06 &0.22 &2.15 &2.41 &6.23 &0.07 &0.14 &0.26 \\
	\bottomrule
	\end{tabular}}
	\label{tab:time}
\end{table*}

\subsection{Comparison with Other Methods}
\subsubsection{Comparison on black-box attacks}
To verify the effectiveness of our approach, we conduct black-box attack experiments on two publicly available datasets, the LADN dataset and CelebA-HQ dataset, by evaluating the attack success rate (ASR) against four different target FR models, \emph{i.e.}, IRSE50, IR152, FaceNet, MobileFace. As shown in Table~\ref{tab:table1}, each column indicates the ASR (\%) results at 0.01 FAR and 0.001 FAR for the specified target model. For each specified target model, the other three remaining models are used as training models. It can be noticed that our approach outperforms the other methods by a large margin and achieves a high attack rate on all four target FR models, indicating that our method yields the best black-box transferability.

Apart from the four face recognition models, we also test on three commercial state-of-the-art FR APIs, \emph{i.e.} Face++, Baidu and Aliyun, to further explore the transferability and reliability of our method.
For each method, we randomly select 1,000 generated adversarial images on each test dataset for evaluations. The results are presented in the box plots in Fig.~\ref{fig:faceapi}. The results show that our method achieves the best average confidence scores for all APIs on each test dataset and our method exceeds average confidence scores of the competitors about 3\%-35\%. Moreover, our method still maintains high confidence scores when facing Aliyun, while other methods have a significant degradation.

\subsubsection{Comparison by simulating real-world scenarios}
To draw near realistic application scenarios, we conduct experiments of face verification and face identification on two large-scale datasets. 

\textbf{Datasets.}
Two large-scale face datasets, namely LFW~\cite{huang2008labeled} and CelebA-HQ~\cite{karras2018progressive}, are employed to assess the performance of our method. (1) Labeled Faces in the Wild (LFW) contains 13,233 images of 5,749 unique subjects and we only consider subjects with at least two face images, resulting in 9,164 images of 1,680 subjects for evaluation. (2) CelebA-HQ dataset comprises 30,000 images of 6,217 unique identities and we similarly filter out subjects with less than two face images, leaving 28,299 images of 4,516 subjects for evaluation.
For both of the test sets, we randomly choose 5 target images for evaluation and report the average results across all target images.

For face verification, we select 10,000 pairs from each dataset, with each pair belonging to the same identity but not the same image. Then, we conduct 10,000$\times$5=50,000 tests for 5 target images. For face identification, we first filter out duplicate identity images from the 10,000 generated images by the face verification section, leaving 5,125 LFW images and 8,951 CelebA-HQ images as the probe sets respectively. Then we use the entire LFW dataset of 13,233 images and the entire CelebA-HQ dataset of 30,000 images as the corresponding gallery sets for evaluation. 
For any image in the probe sets, there is at least one image in the gallery sets that belongs to the same identity as the image. This is because face encryption libraries in social media typically include multiple images of each identity and hide multiple images for each identity is more difficult but more practical. Similarly, we conduct 5,125$\times$5=25,625 tests for LFW and 8,951$\times$5=44,755 tests for CelebA-HQ.

\textbf{Evaluation Metrics.}
For face verification, we report the \emph{protection success rate} (PSR, higher is better), which is the fraction of generated images that are not classified to the original source images according to the $\tau$ threshold. It can be computed by
\begin{equation}
	PSR=\frac{\sum_{i=1}^{N} 1_{\tau}\left(\cos \left[H\left(I_{t}^{i}\right), H\left({I}_{src}\right)\right]<\tau\right)}{N} \times 100 \%\text{,}
\label{eq:psr}
\end{equation}
where H($\cdot$) represents the target FR model, $I_{src}^i$ is the $i$-th source identity, and $I_t^i$ represents the generated $i$-th adversarial example. $1_{\tau}$ is the indicator function of protection success, where its value is 1 if the feature distance between two face images is less than the pre-determined threshold, indicating that the images are assumed to not belong to the same person. Here, the values of $\tau$ for all face recognition models are set as the thresholds at 0.01 and 0.001 FAR (False Acceptance Rate). 

For face identification, we use the Rank-1 and Rank-5 protection success rates (higher is better) as metrics. These rates measure the proportion of probe images whose closest 1 or 5 neighbors in the gallery set do not belong to the same identity. Specifically, for each probe image, we calculate the distance between the probe image and each image in the gallery set and rank them accordingly. For Rank-1 and Rank-5 protection success, the top-1 and top-5 closest images should not have the same identity as the probe image. The success rate is then calculated as the number of images that are protected successfully.

\textbf{Results.} 
For face verification, it is commonly utilized in real-world scenarios like identity verification and face payment. To verify the identity protection performance of our method in such contexts, we evaluate the protection success rate (PSR) under the face verification setting on two large-scale face recognition datasets, LFW dataset and CelebA-HQ dataset. As shown in Table~\ref{tab:table_pro}, we report the PSR results for four different target models and different FARs, with each column representing the PSR results of the specific target model at 0.01 FAR and 0.001 FAR. It can be noticed that the protection success rate of clean images, \emph{i.e.}, original data without any protection, is very low. This means that if unauthorized face recognition systems are exposed to original images with the same identity as the source image, the privacy and security of personal information may be significantly threatened. Differently, our model achieves the highest protection success rate under different target models and FAR thresholds, demonstrating the highest potential for real-world identity protection.

Unlike face verification, face identification is frequently employed in real-world applications such as face attendance and surveillance systems. To evaluate the identity concealment of our method in these scenarios, we report the Rank-1 and Rank-5 protection success rates under the face identification setting. Table~\ref{tab:table_rank} presents our results for four different target models and two large-scale datasets, with higher scores indicating better performance. Our method continues to outperform other methods in the face identification setting, demonstrating its effectiveness and generalizability.

\subsubsection{Comparison on image quality}
In the following, we show quantitative and qualitative results of our method to verify its visual quality. As shown in Table~\ref{tab:table2}. our method achieves satisfying performance in terms of FID, SSIM, and PSNR. 

Compared with AMT-GAN, our method achieves better FID results, which indicates that our method can generate more natural and imperceptible adversarial facial images. We also compare our method with the state-of-the-art methods related to makeup transfer, including PSGAN~\cite{jiang2020psgan}, PSGAN++~\cite{liu2021psgan++}, LADN~\cite{gu2019ladn}, and SOGAN~\cite{lyu2021sogan}. Our method achieves comparable performance in the aspect of makeup transfer, even surpassing LADN and SOGAN in terms of the FID metric. It is noted that Adv-Makeup achieves the best results in all quantitative metrics in Table~\ref{tab:table2}. It is because Adv-Makeup only produces adversarial eye shadow over a small orbital region on faces, making it difficult to protect the source facial images against strong FR models, as shown in Fig.~\ref{fig:faceapi} and Fig.~\ref{fig:quali}. For example, Adv-Makeup has a low attack success rate of 11.68\% on the CelebA-HQ dataset when attacking the FaceNet model.

In Fig.~\ref{fig:quali}, we show a qualitative comparison with methods that construct adversarial images based on makeup transfer, \emph{i.e.}, Adv-Makeup and AMT-GAN. The three numbers below each face image are the confidence scores produced by the commercial online FR systems, Face++, Baidu and Aliyun. Obviously, the adversarial images generated by Adv-Makeup have sharp boundaries around the eye regions as it synthesizes adversarial eye shadows in a patch-based manner. 
AMT-GAN can generate more natural adversarial faces than Adv-Makeup. However, AMT-GAN performs not well in accurately transferring the makeup styles from the reference images to the source images, as shown in the third column of Fig.~\ref{fig:quali}. Compared with the above methods based on makeup transfer, our method not only performs makeup transfer accurately, but also achieves better visual quality and black-box transferability on commercial FR APIs. 

\begin{figure}[tp]
	\centering
		\includegraphics[width=1\linewidth]{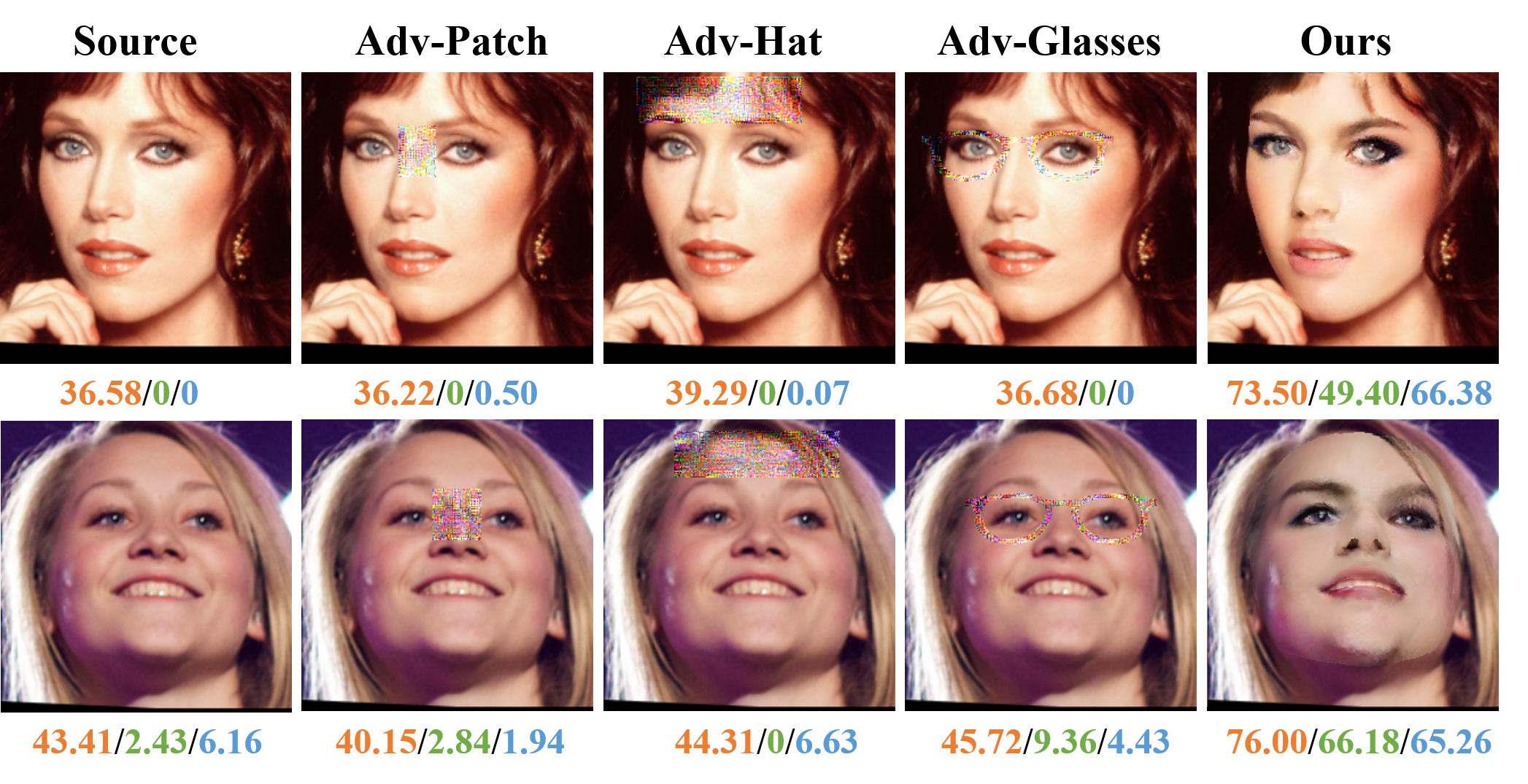}
		\caption{Comparison with patch-based methods, \emph{i.e.}, Adv-Patch, Adv-Hat and Adv-Glasses. The three numbers below each face image are the confidence scores returned by the commercial online FR APIs, Face++, Baidu and Aliyun.}
	\label{fig:patch}
\end{figure}

In addition, we give a qualitative comparison with patch-based methods, Adv-patch, Adv-Hat and Adv-Glasses. As shown in Fig.~\ref{fig:patch}, the adversarial patches generated by Adv-patch, Adv-Hat and Adv-Glasses have large size of adversarial areas on the faces that are easily noticed by human eyes. Differently, our method can generate the protected faces indistinguishable from the corresponding original images.

\subsubsection{Comparison on Time Efficiency}
As illustrated in Table~\ref{tab:time}, we compare the computation time with other approaches. Most gradient-based approaches, such as PGD~\cite{kurakin2017adversarial}, MI-FGSM~\cite{dong2018boosting}, TI-DIM~\cite{dong2019evading} and TIP-IM~\cite{yang2021towards}, generate adversarial images with an iterative process. In our experiments, we set the number of iteration steps to 10 to maintain a trade-off between the efficiency and effectiveness. However, these approaches still need several seconds during inference. The gradient-based method, FGSM~\cite{dong2019evading}, is fast due to the single-step setting, but it does not have a strong ASR performance, as shown in Table~\ref{tab:table1}. Compared with these gradient-based approaches, the methods based on makeup transfer are about 8-40 times faster.

\begin{figure}[tp]
	\centering
		\includegraphics[width=1\linewidth]{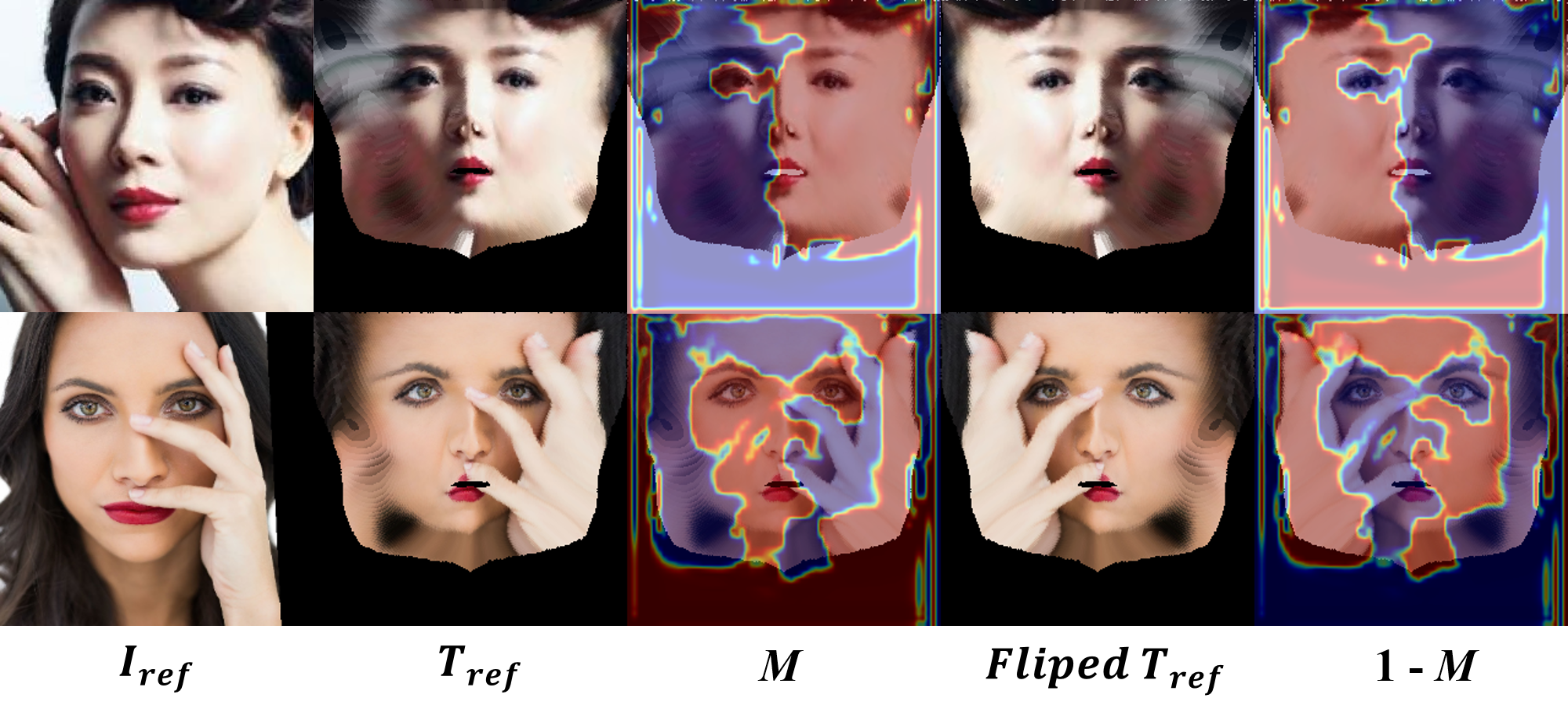}
		\caption{Visualizations of the attentional mask $M$ in MAM. $M$ learns the makeup quality of different regions and gives higher confidence to cleaner regions and lower confidence to contaminated regions.}
	\label{fig:visual}
\end{figure}

\begin{figure}[tp]
	\centering
		\includegraphics[width=1\linewidth]{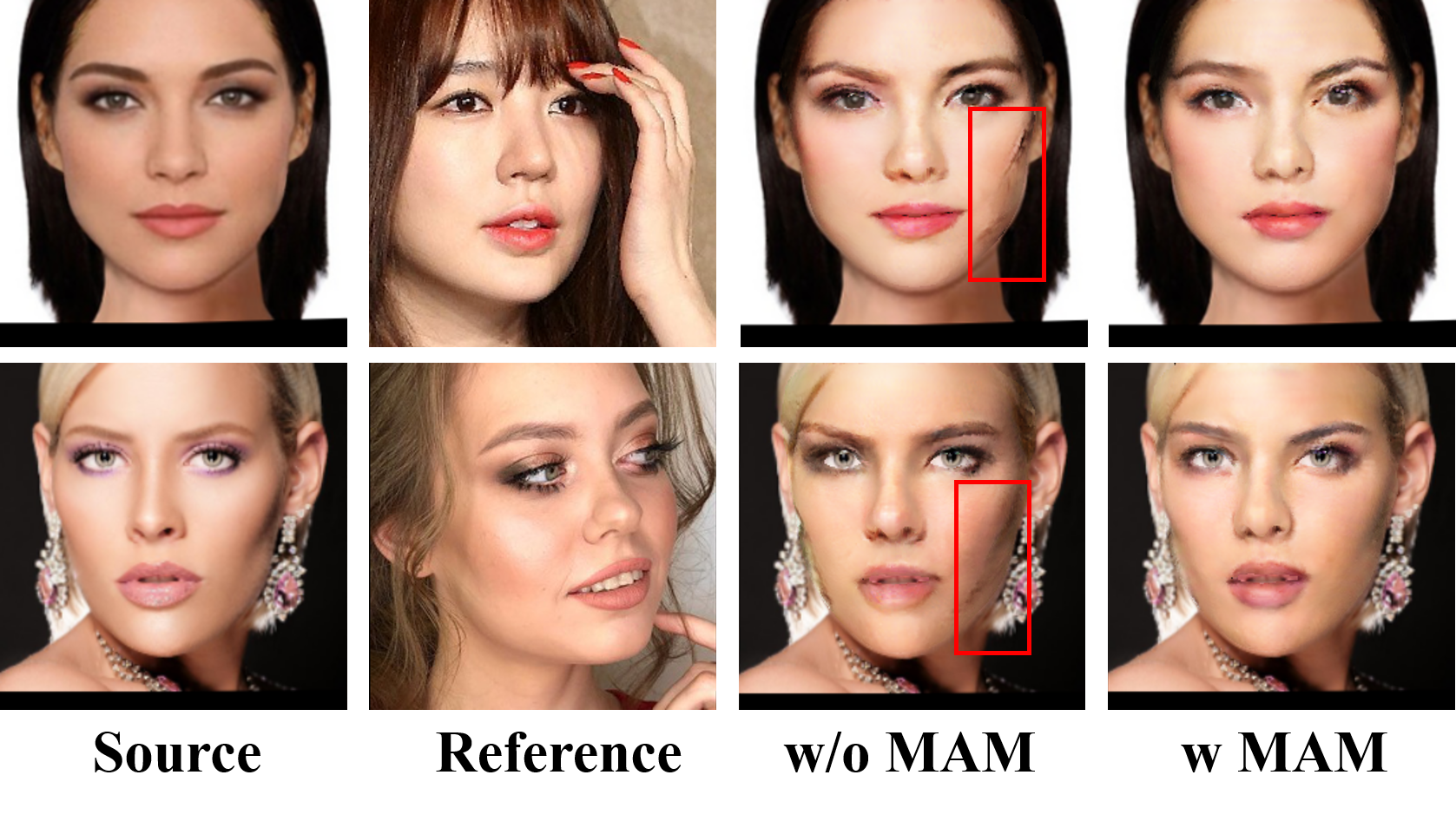}
		\caption{Ablation study of the proposed MAM module.}
	\label{fig:womam}
\end{figure}

\begin{figure}[t!]
	\centering
		\includegraphics[width=1\linewidth]{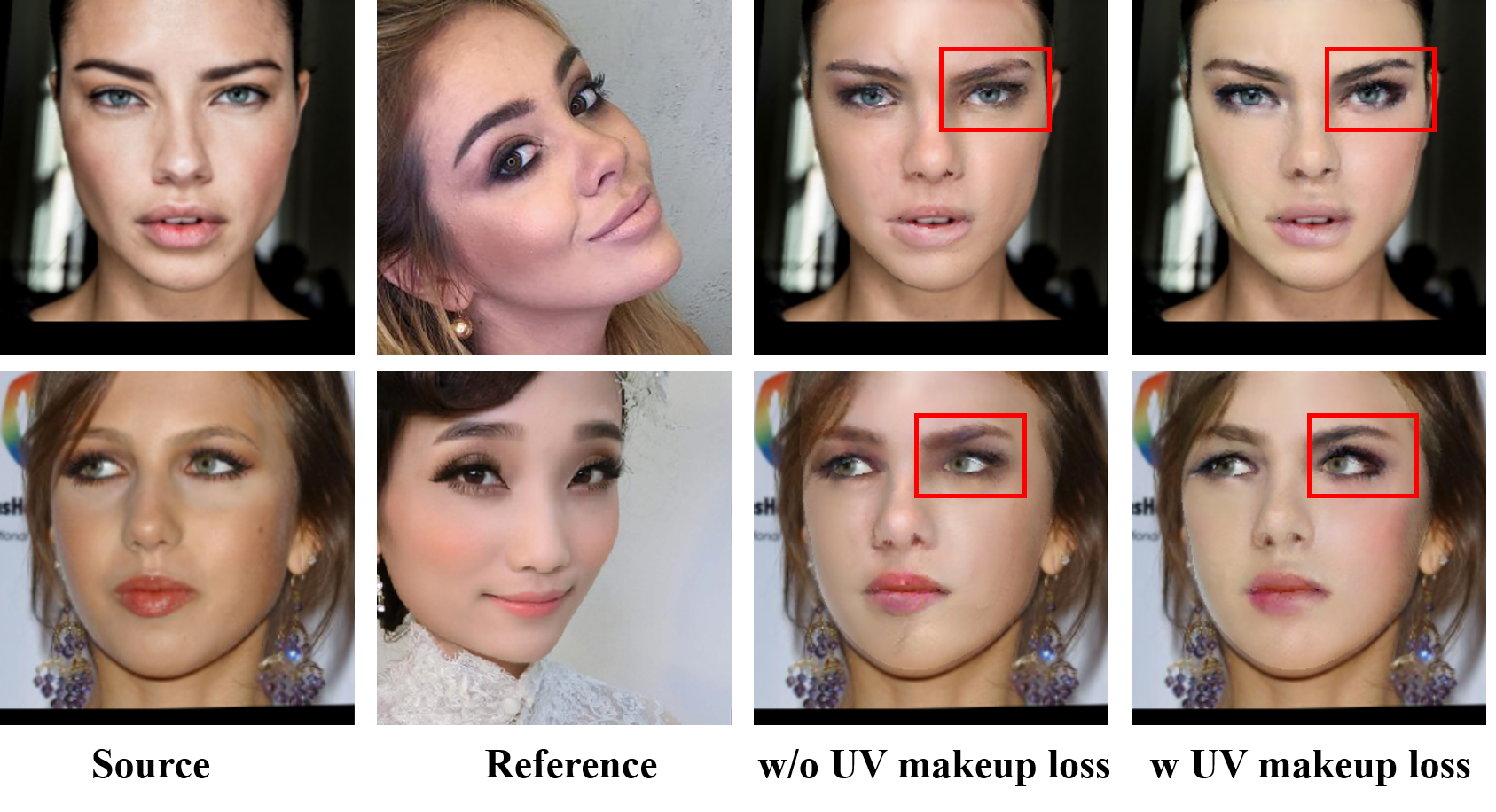}
		\caption{Ablation study of the proposed UV makeup loss.}
	\label{fig:woloss}
\end{figure}

\begin{figure*}[tp]
	\centering
		\includegraphics[width=1\linewidth]{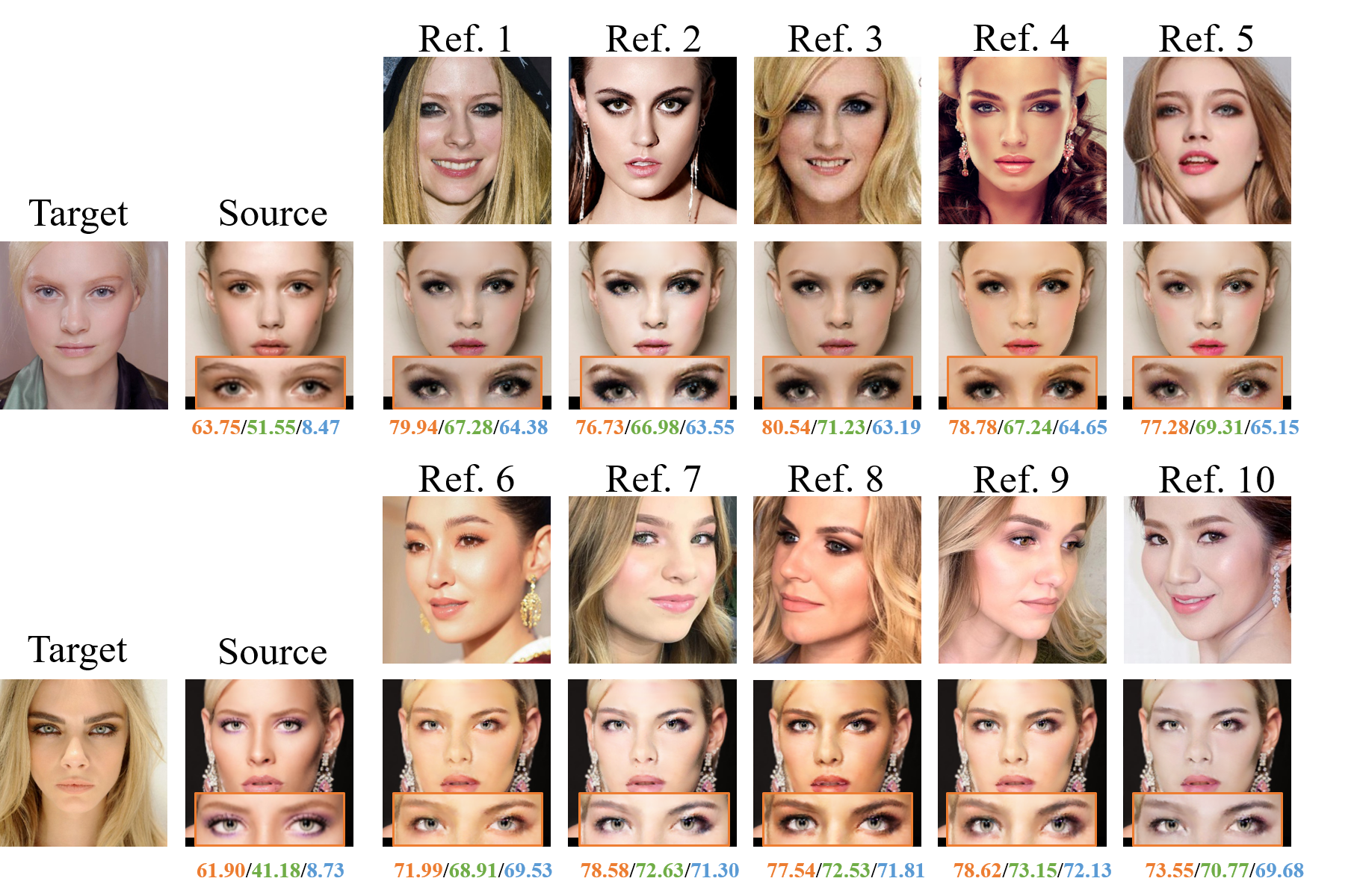}
		\caption{Qualitative evaluations of the impact of different makeup styles. The three numbers below each face image are the confidence scores returned by the commercial online FR APIs, Face++, Baidu and Aliyun. Although the selected reference makeups are different in styles, poses and expressions, our method can perform accurate and robust makeup transfer and maintain a high black-box performance on three FR APIs.}
	\label{fig:ref}
\end{figure*}
\begin{figure}[tp]
	\centering
		\includegraphics[width=1\linewidth]{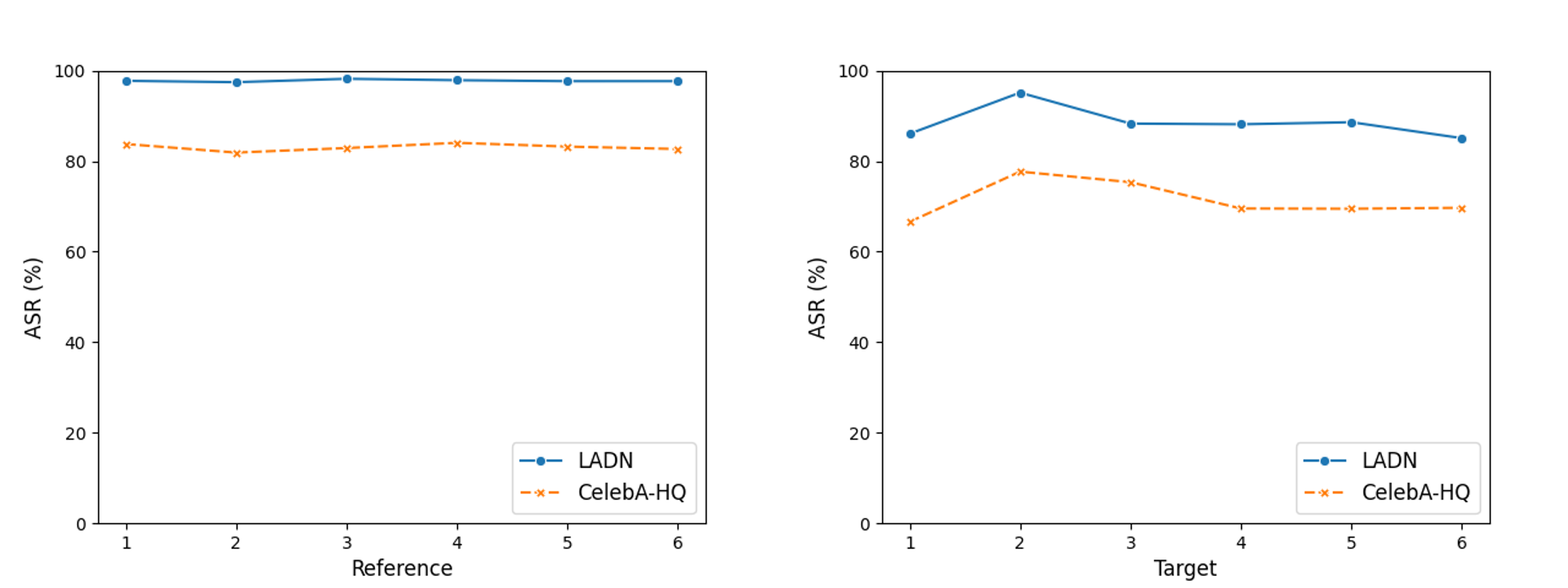}
		\caption{Evaluations of the robustness under different makeup styles and different target identities. We show the average ASR~(\%) results for different reference images and different target images on the LADN dataset and CelebA-HQ dataset.}
	\label{fig:tarref}
\end{figure}

\subsection{Ablation Studies}

\subsubsection{Effectiveness of Makeup Adjustment Module (MAM)}
The makeup quality of chosen reference images influences the robustness of the adversarial makeup generation. Thus, we propose the Makeup Adjustment Module (MAM) to locate, catch and repair the imperfect makeup regions and improve the visual quality of the generated images. In the following, we verify the effectiveness of the proposed MAM. As shown in Fig.~\ref{fig:visual}, the first two columns are the input reference images and the corresponding UV texture maps, they have the shadow and occlusion caused by the pose or self-occlusion, which will cause ghosting artifacts in the generated images. By applying MAM with a 3D visibility map as the additional input, the unclean regions can be located by the attentional mask $M$. For example, in the first row, the attentional mask $M$ gives high confidence to the clean regions and low confidence to the shadowed area on the left-side face. In the second row, for the occlusion caused by the hand, the attentional mask $M$ also gives accurate confidence. Guided by $M$, MAM can use information from both the original and the flipped texture map to repair the inaccurate regions by Eq.~(\ref{con:MAM}).

Furthermore, in Fig.~\ref{fig:womam}, we show the qualitative results by removing the MAM module. The first column and the second column are the input source images and reference images, respectively. As a result, the transferred image shows occlusion artifacts (marked in the red boxes) in the same regions without MAM, which is caused by the shadowed makeup in the reference image. 
By introducing MAM, it can adaptively fix the imperfect reference makeup.

Finally, to further verify the effectiveness of the proposed MAM, we give the quantitative comparison between ``w / o MAM'' and ``Ours''. As shown in Table~\ref{tab:table2}, our method outperforms ``w / o MAM'' on all evaluation metrics, FID, SSIM and PSNR, which indicates the proposed MAM module can effectively improve the visual quality of makeup generation.

\subsubsection{Effectiveness of UV makeup loss}
To further improve the visual quality of the adversarial makeup generation, we introduce a UV makeup loss by leveraging the bilateral symmetry of faces in the UV space. In this section, we conduct ablation studies for evaluating the effectiveness of the proposed loss function. As shown in Table~\ref{tab:table2}, compared with the results of ``w / o UV makeup loss'', our method show better performance in terms of all quantitative metrics, especially the FID metric. This indicates that our method with the proposed loss can better handle reference images with large poses and achieve more robust and accurate makeup generation. As shown in Fig.~\ref{fig:woloss}, the first column and the second column are the input source images and reference images, respectively. When the reference makeup images have occlusion caused by the face pose, the method without the UV makeup loss can not accurately grasp the corresponding makeup information in the occluded makeup regions (shown in the third column). By introducing the UV makeup loss, the method can low the undesired effect caused by large pose and derive more accurate and robust results (shown in the forth column).

\subsubsection{Robustness on different references and targets}
We show the robustness of our approach on different references and targets. As shown in Fig.~\ref{fig:ref}, for each target image, we randomly choose five reference images~(Ref. 1-Ref. 5) from the MT-dataset~\cite{li2018beautygan} and five reference images~(Ref. 6-Ref. 10) from the Makeup-Wild (MW) dataset~\cite{jiang2020psgan} for evaluations, where the MW dataset is a collected dataset that contains images with large poses and expressions. The three numbers below each face image are the confidence scores returned by the commercial online FR systems, Face++, Baidu and Aliyun. It can be observed that although the given reference images have different makeup styles, different poses and expressions, our approach can accurately transfer the fine-grained makeup styles to the source images with high black-box transferability for all the given target images, which demonstrates the effectiveness and robustness of our approach.

In addition, we further test the average ASR(\%) results for different reference images and different target images on the LADN dataset and CelebA-HQ dataset in Fig.~\ref{fig:tarref}. 
In the left figure of Fig.~\ref{fig:tarref}, the changes of different reference makeup styles have very little effect on the ASR performance. In the right figure of Fig.~\ref{fig:tarref}, the changes in different target images have a noticeable effect on the ASR performance. Nevertheless, the results under different conditions (\emph{i.e.}, different references or targets) are both stable and maintain a high average ASR performance.

\subsection{Physical-world experiments}
In this section, we demonstrate the protected images generated by our method are physically realizable and robust, as well as their persistent superiority after printing and re-photographing. Specifically, we randomly select 5 pairs of images with the same identity from the CelebA-HQ dataset for four face recognition models. For each pair, we apply adversarial makeup to one image to conceal the identity, print and re-photograph it, and employ the other image for face verification. We compare our method against different benchmark schemes of adversarial attacks, including gradient-based TI-DIM~\cite{dong2019evading} and TIP-IM~\cite{yang2021towards}, as well as adversarial makeup-based Adv-makeup~\cite{yin2021adv} and AMTGAN~\cite{hu2022protecting}. We measure the cosine distance between the protected image after printing and re-photographing and the other image with the same identity, where larger distances indicate stronger protection.
As shown in Table~\ref{tab:tab_phy}, the average cosine similarity of the original images (clean images) after re-photographing still exceeds the threshold at 0.01 FAR for different FR models (\emph{i.e.}, 0.241 for IRSE50, 0.167 for IR152, 0.302 for MobileFace, 0.409 for FaceNet), suggesting that the original results can still be recognized as the same identity. All methods can reduce the cosine distance between the two images. Notably, our method obtains the minimum average distance between the protected image and the other image with the same identity, highlighting its high identity protection performance even after printing and re-photographing.
 
\begin{table}[htb]
	\caption{The cosine similarities of different methods between the protected images (after printing and re-photographing) and the images with the same identity.}
	\centering
	\label{tab:tab_phy}
	\resizebox{0.9\columnwidth}{!}{
	\begin{tabular}{c|llll}
	\hline
	\multicolumn{1}{l|}{} & \multicolumn{1}{c}{IRSE50} &\multicolumn{1}{c}{IR152} &\multicolumn{1}{c}{FaceNet} &\multicolumn{1}{c}{MobileFace} \\ \hline
	Clean                 & \multicolumn{1}{c}{0.67$\pm$0.06} & 0.35$\pm$0.13  & 0.68$\pm$0.10& 0.50$\pm$0.12  \\ \hline
	TI-DIM~\cite{dong2019evading} & 0.63$\pm$0.08      & 0.26$\pm$0.15  & 0.64$\pm$0.14 & 0.46$\pm$0.10       \\
	TIP-IM~\cite{yang2021towards} & 0.55$\pm$0.06      & 0.24$\pm$0.19  & 0.56$\pm$0.17 & 0.44$\pm$0.13       \\
	Adv-makeup~\cite{yin2021adv} & 0.46$\pm$0.15  & 0.26$\pm$0.14  & 0.58$\pm$0.16    & 0.36$\pm$0.10     \\
	AMTGAN~\cite{hu2022protecting} & 0.36$\pm$0.09 & 0.11$\pm$0.09  & 0.48$\pm$0.08    & 0.27$\pm$0.14       \\ \hline
	Ours   & \textbf{0.34$\pm$0.07} & \textbf{0.07$\pm$0.09}  & \textbf{0.29$\pm$0.10} & \textbf{0.27$\pm$0.12} \\ \hline
	\end{tabular}}
	\end{table}

\section{Conclusion and Future Work}
In this paper, we propose a 3D-Aware Adversarial Makeup Generation GAN (3DAM-GAN). 3DAM-GAN constructs adversarial examples based on makeup transfer, which is a powerful and natural approach for facial privacy protection.
It contains a UV generator to transfer the makeup in the UV space and a makeup attack mechanism to make the generated makeup adversarial. Specifically, the UV generator consists of the proposed Makeup Adjustment Module (MAM) and Makeup Transfer Module (MTM), which work together to improve the fidelity of the synthetic makeup. In addition, to further provide more accurate makeup supervision, a UV makeup loss is proposed by leveraging the inherent symmetry of human faces in UV space.
To encourage the generator with high black-box transferability, we introduce a makeup attack mechanism with ensemble training strategy to learn common and informative gradient directions against various FR models. Finally, the source faces with adversarial makeup are protected by optimizing the source identity to another arbitrarily selected target identity. Extensive experiments on multiple datasets and target models demonstrate our approach has strong identity protection ability (black-box transferability) and high visual quality (naturalness).

\textbf{Limitation.} While 3DAM-GAN shows promising results in our experiments, there are some limitation to the proposed approach is that it may not be suitable for individuals who do not wish to be presented with makeup, such as many men in certain cultures.

As for future directions, we plan to develop a more balanced dataset with makeup styles since the current training dataset (MT-dataset) is unbalanced in gender, resulting in better visual quality of female images. Besides, we plan to explore other techniques to protect facial privacy that may be more universally acceptable, \emph{e.g.}, using relighting techniques to construct adversarial perturbations. 

\section{Acknowledgements} 
This work is supported by the National Key Research and Development Program of China under Grant No. 2021YFC3320103, the National Natural Science Foundation of China (NSFC) under Grant 62272460, the Young Elite Scientists Sponsorship Program by CAST (YESS), and the CAAI-Huawei MindSpore Open Fund.

\ifCLASSOPTIONcaptionsoff
  \newpage
\fi

\bibliographystyle{IEEEtran}
\bibliography{main}
%
\begin{IEEEbiography}[{\includegraphics[width=1in,height=1.25in,clip,keepaspectratio]{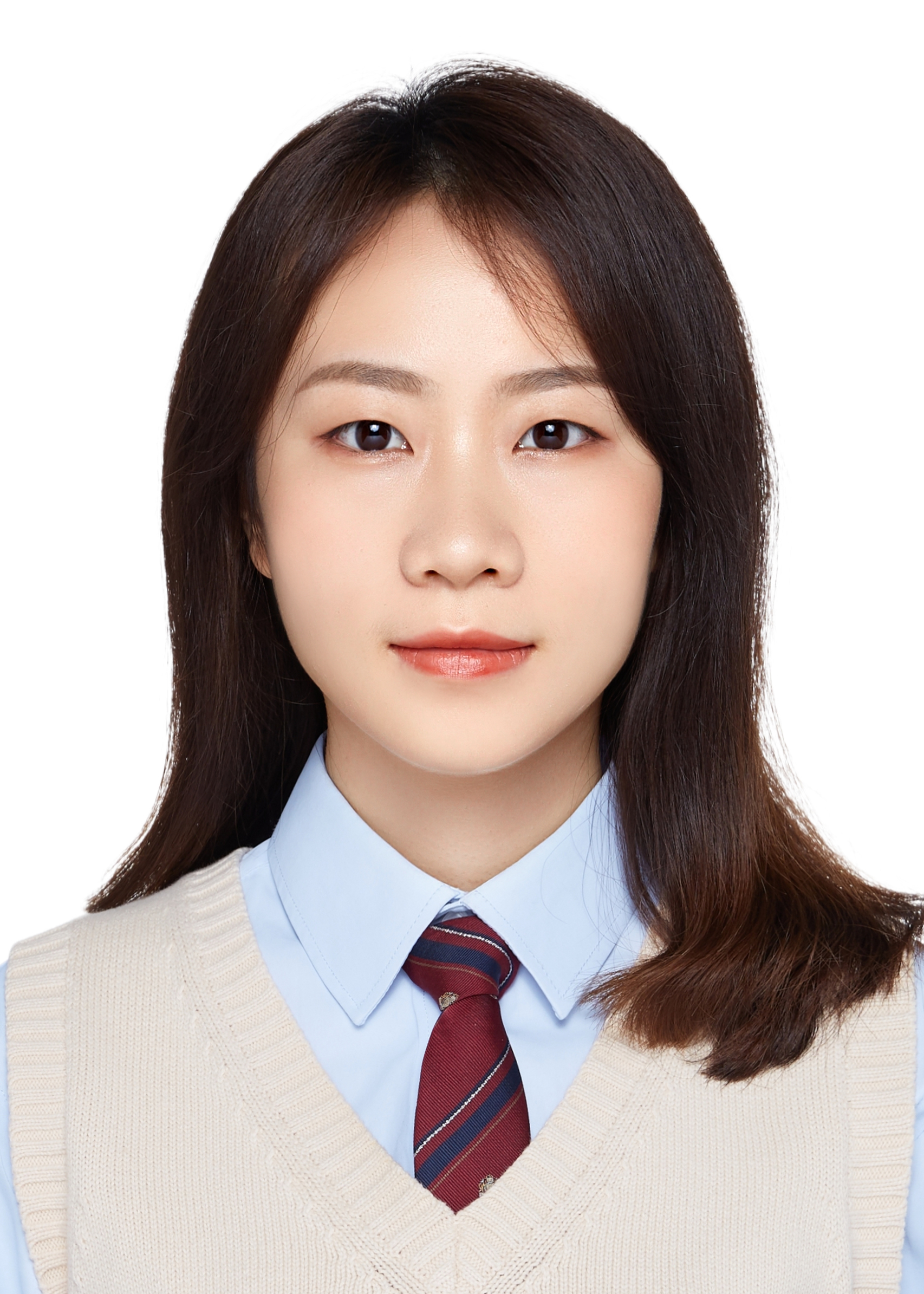}}]{Yueming Lyu}{\space}received B.Eng. degree in Nanjing University of Aeronautics and Astronautics, China in 2019. She is a Ph.D. degree candidate in the Center for Research on Intelligent Perception and Computing (CRIPAC) at the State Key Laboratory of Multimodal Artificial Intelligence Systems, Institute of Automation, Chinese Academy of Sciences, China. Her current research include computer vision, image generation and adversarial learning.

\end{IEEEbiography}

\begin{IEEEbiography}[{\includegraphics[width=1in,height=1.25in,clip,keepaspectratio]{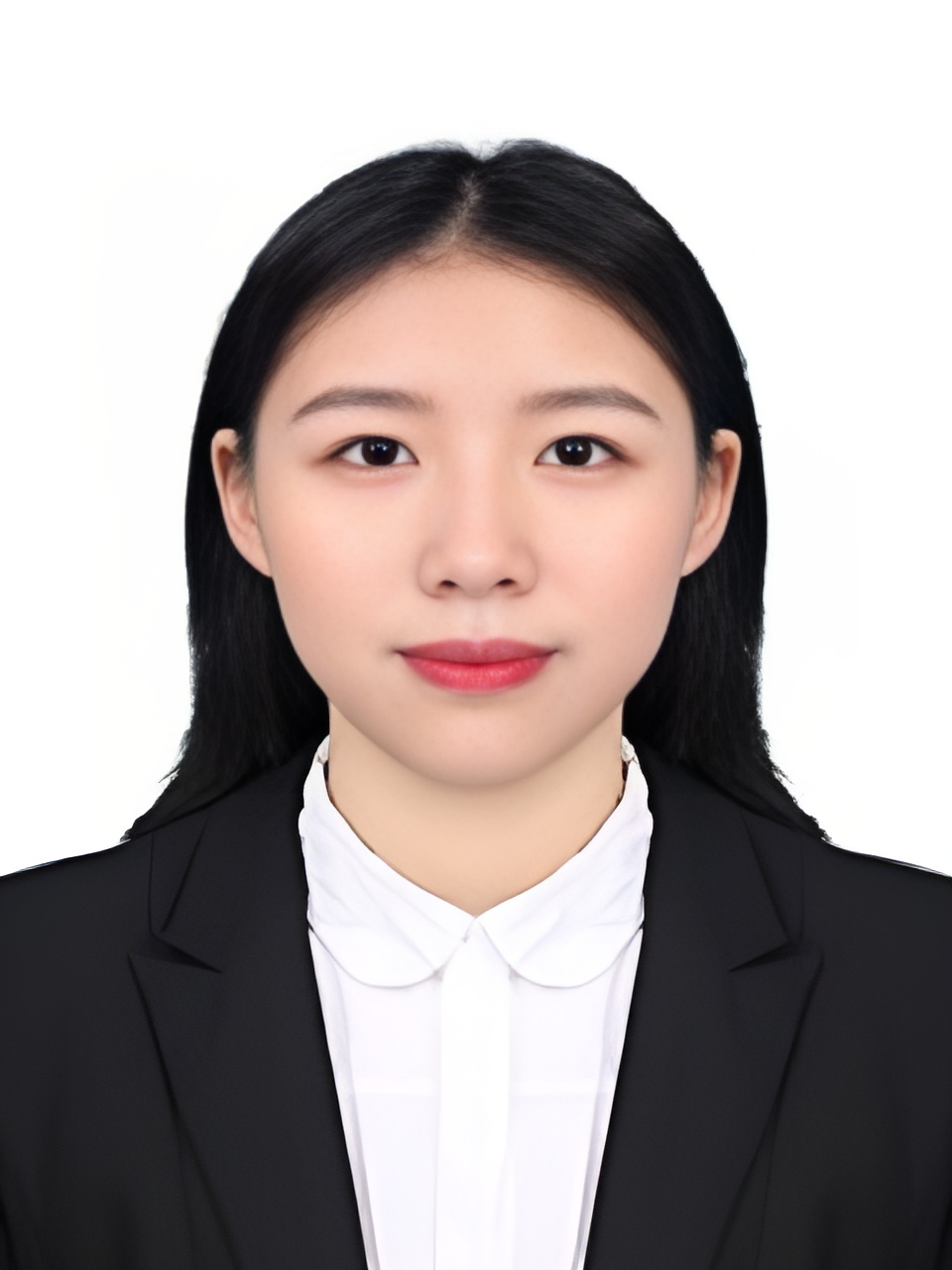}}]{Yue Jiang}{\space}received B.Eng. degree in Xi'an Jiaotong University, China in 2021. She is a Master degree candidate in the Center for Research on Intelligent Perception and Computing (CRIPAC) at the State Key Laboratory of Multimodal Artificial Intelligence Systems, Institute of Automation, Chinese Academy of Sciences, China. Her current research focuses on computer vision and image generation.

\end{IEEEbiography}

\begin{IEEEbiography}[{\includegraphics[width=1in,height=1.25in,clip,keepaspectratio]{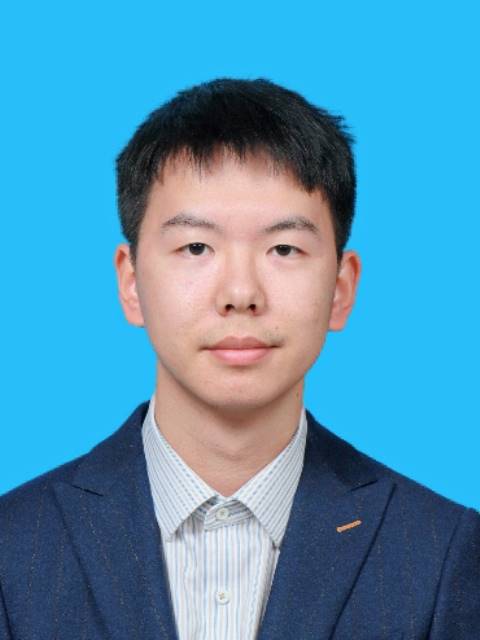}}]{Ziwen He}{\space}received B.Eng. degree in Shanghai Jiao Tong University, China in 2018. He is a Ph.D. degree candidate in the Center for Research on Intelligent Perception and Computing (CRIPAC) at the State Key Laboratory of Multimodal Artificial Intelligence Systems, Institute of Automation, Chinese Academy of Sciences, China. His current research focuses on adversarial example and computer vision.

\end{IEEEbiography}

\begin{IEEEbiography}[{\includegraphics[width=1in,height=1.25in,clip,keepaspectratio]{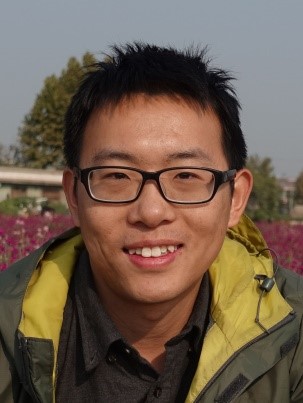}}]{Bo Peng}{\space}received B.Eng. degree in Beihang University and PhD degree in the Institute of Automation, Chinese Academy of Sciences in 2013 and 2018, respectively. Since July 2018, Dr. Bo Peng has joined the Institute of Automation, Chinese Academy of Sciences where he is currently an Associate Professor. His current research focuses on computer vision and image forensics. 

\end{IEEEbiography}

\begin{IEEEbiography}[{\includegraphics[width=1in,height=1.25in,clip,keepaspectratio]{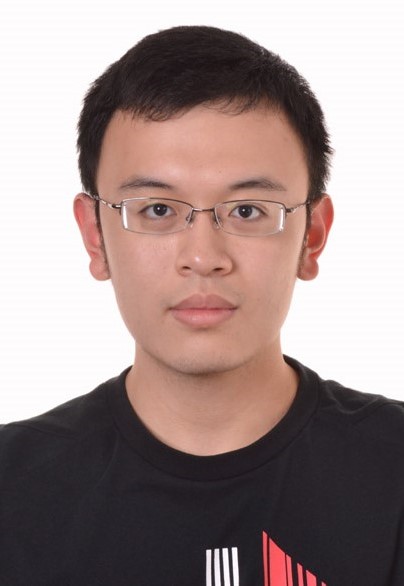}}]{Yunfan Liu}{\space}received the B.E. degree in electronic engineering from Tsinghua University, Beijing, China, in 2015, the M.S. degree in electronic engineering: systems from University of Michigan, Ann Arbor, United States, in 2017. He is currently pursuing the Ph.D. degree with the Center for Research on Intelligent Perception and Computing, State Key Laboratory of Multimodal Artificial Intelligence Systems, Institute of Automation, Chinese Academy of Sciences (CASIA), Beijing, China. His research interests include omputer vision, pattern recognition, and machine learning.

\end{IEEEbiography}

\begin{IEEEbiography}[{\includegraphics[width=1in,height=1.25in,clip,keepaspectratio]{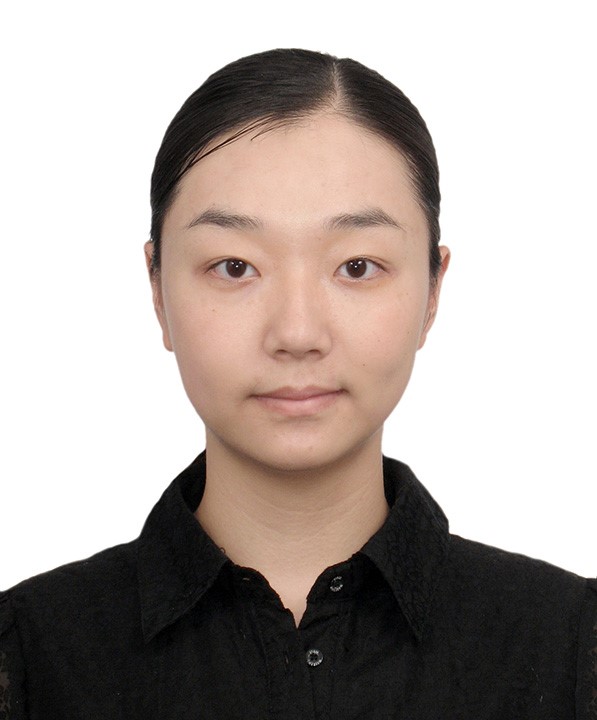}}]{Jing Dong}{\space}recieved her Ph.D in Pattern Recognition from the Institute of Automation, Chinese Academy of Sciences, China in 2010. Then she joined the Institute of Automation, Chinese Academy of Sciences and she is currently a Professor. Her research interests are towards Pattern Recognition, Image Processing and Digital Image Forensics including digital watermarking, steganalysis and tampering detection. She is a senior member of IEEE. She also has served as the deputy general of Chinese Association for Artificial Intelligence.

\end{IEEEbiography}
\vfill









\end{document}